\documentclass[10pt,journal,compsoc]{IEEEtran}
\ifCLASSOPTIONcompsoc
  \usepackage[nocompress]{cite}
\else
  \usepackage{cite}
\fi
\ifCLASSINFOpdf
\else
\fi

\usepackage{subfigure,multirow,array,graphicx,amssymb,amsmath}

\usepackage{ragged2e}

\newcommand{\tabincell}[2]{\begin{tabular}{@{}#1@{}}#2\end{tabular}}

\newcommand{\mathbbm}[1]{\text{\usefont{U}{bbm}{m}{n}#1}}

\begin{document}
%
\title{Unconstrained Facial Action Unit Detection via Latent Feature Domain}
%
%
%
%

\author{Zhiwen~Shao,
        Jianfei~Cai,~\IEEEmembership{Fellow,~IEEE,}
        Tat-Jen~Cham,
        Xuequan~Lu,
        and~Lizhuang~Ma
\IEEEcompsocitemizethanks{
\IEEEcompsocthanksitem Z. Shao is with the School of Computer Science and Technology, China University of Mining and Technology, Xuzhou 221116, China, and also with the Engineering Research Center of Mine Digitization, Ministry of Education of the People’s Republic of China, Xuzhou 221116, China. E-mail: zhiwen\_shao@cumt.edu.cn.
\IEEEcompsocthanksitem J. Cai is with the Faculty of Information Technology, Monash University, Victoria 3800, Australia. E-mail: jianfei.cai@monash.edu.
\IEEEcompsocthanksitem T.-J. Cham is with the School of Computer Science and Engineering, Nanyang Technological University, Singapore 639798. E-mail: astjcham@ntu.edu.sg.
\IEEEcompsocthanksitem X. Lu is with the School of Information Technology, Deakin University, Victoria 3216, Australia. E-mail: xuequan.lu@deakin.edu.au.
\IEEEcompsocthanksitem L. Ma is with the Department of Computer Science and Engineering, Shanghai Jiao Tong University, Shanghai 200240, China, also with the MoE Key Lab of Artificial Intelligence, Shanghai Jiao Tong University, Shanghai 200240, China, and also with the School of Computer Science and Technology, East China Normal University, Shanghai 200062, China. E-mail: ma-lz@cs.sjtu.edu.cn.
}
\thanks{Manuscript received June, 2020. (Corresponding authors: Zhiwen~Shao and Lizhuang~Ma.)}}

%
%

\markboth{IEEE Transactions on Affective Computing,~Vol.~X,~NO.~X,~X}%
{Shell \MakeLowercase{\textit{et al.}}: Bare Demo of IEEEtran.cls for Computer Society Journals}
%



\IEEEtitleabstractindextext{%
\begin{abstract}
\justifying Facial action unit (AU) detection in the wild is a challenging problem, due to the unconstrained variability in facial appearances and the lack of accurate annotations. Most existing methods depend on either impractical labor-intensive labeling or inaccurate pseudo labels. In this paper, we propose an end-to-end unconstrained facial AU detection framework based on domain adaptation, which transfers accurate AU labels from a constrained source domain to an unconstrained target domain by exploiting labels of AU-related facial landmarks. Specifically, we map a source image with label and a target image without label into a latent feature domain by combining source landmark-related feature with target landmark-free feature. Due to the combination of source AU-related information and target AU-free information, the latent feature domain with transferred source label can be learned by maximizing the target-domain AU detection performance. Moreover, we introduce a novel landmark adversarial loss to disentangle the landmark-free feature from the landmark-related feature by treating the adversarial learning as a multi-player minimax game. Our framework can also be naturally extended for use with target-domain pseudo AU labels. Extensive experiments show that our method soundly outperforms lower-bounds and upper-bounds of the basic model, as well as state-of-the-art approaches on the challenging in-the-wild benchmarks. The code is available at \textit{https://github.com/ZhiwenShao/ADLD}.
\end{abstract}

}

\maketitle

\IEEEdisplaynontitleabstractindextext

%
\IEEEpeerreviewmaketitle

\section{Introduction}
\label{sec:introduction}

Facial action unit (AU) detection~\cite{zhao2016deep,li2018eac,shao2018deep,corneanu2018deep,niu2019local,ma2019r,jyoti2019expression} involves determining the presence of each AU in a given face image. It has gained increasing attention in computer vision and affective computing communities, due to the use of identifying human emotions in various applications. Each AU is a basic facial action for describing facial expressions, as defined by the Facial Action Coding System (FACS)~\cite{ekman1978facial,ekman2002facial}. While AU detection for near-frontal faces in constrained laboratory conditions~\cite{lucey2010extended,mavadati2013disfa,zhang2014bp4d} has achieved remarkable success, AU detection in the wild~\cite{benitez2016emotionet} still remains a challenge. Compared with images captured under fixed conditions, unconstrained images exhibit a wide variability in expressions, poses, ages, illumination, accessories, occlusions, backgrounds and image quality. Furthermore, due to a limited number of experts and the labor-intensive work required~\cite{TAC2017Pantic}, it is costly and impractical to manually annotate unconstrained images at a large scale for fully-supervised learning.

\begin{figure}
\centering\includegraphics[width=\linewidth]{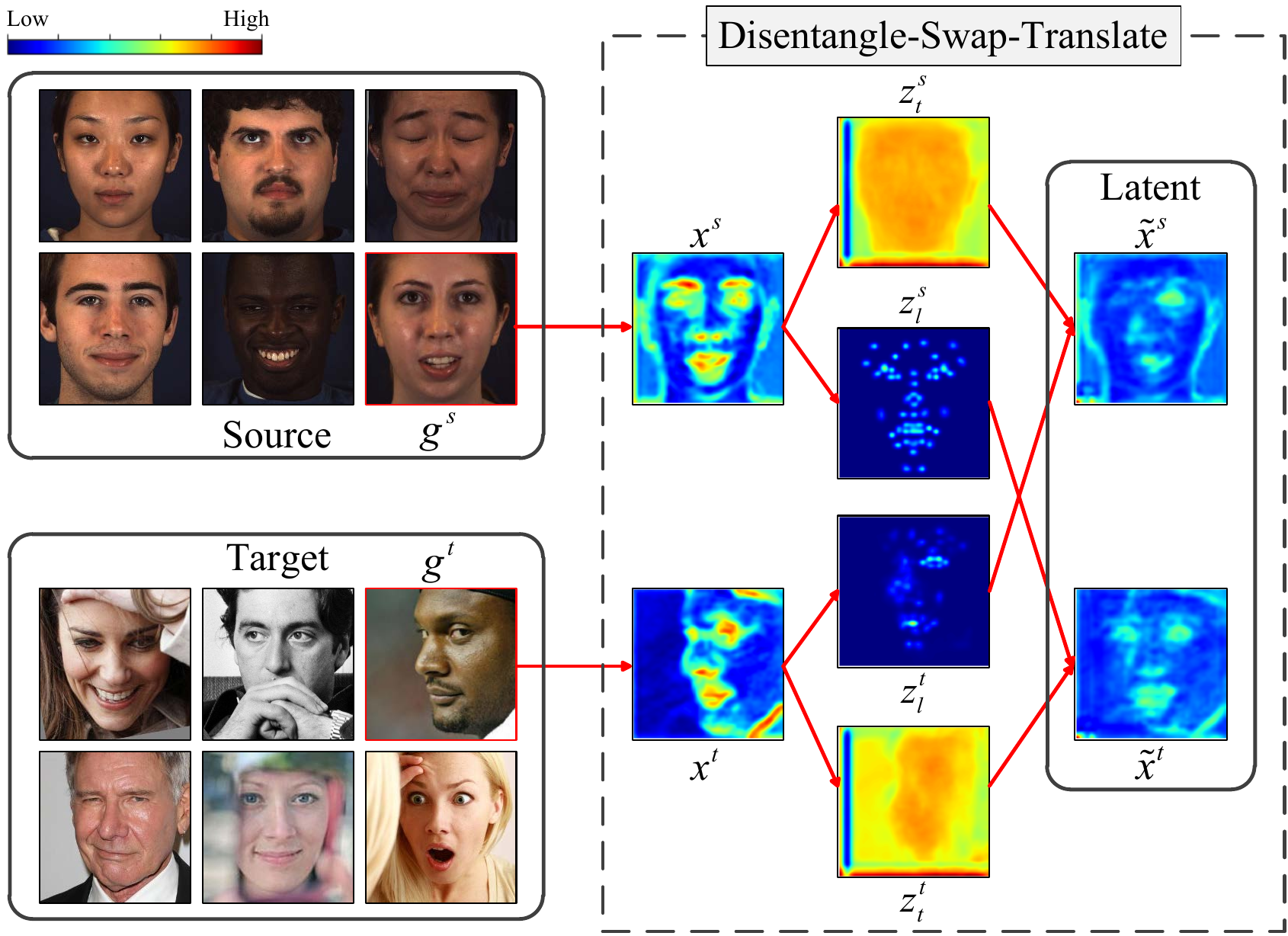}
\caption{Illustration of mapping an image $g^s$ in the source domain and an image $g^t$ in the target domain into the latent feature domain. \textit{The rich features $(x^s, x^t)$ are first disentangled into landmark-free features $(z^s_{t}, z^t_{t})$ and landmark-related features $(z^s_{l}, z^t_{l})$, and then the landmark-related features are swapped to generate latent features $(\tilde{x}^s, \tilde{x}^t)$. The latent feature domain is specialized for target-domain AU detection.} The channels of features are summed element-wise for visualization, where the colors from blue to red in the color bar indicate rising feature values.}
\label{fig:disentangle}
\end{figure}

\noindent\textbf{Limitations of Existing Solutions.} There have been some attempts at AU detection of unconstrained images, which often depend on pseudo AU labels. These pseudo labels were automatically annotated by an AU detection model~\cite{benitez2016emotionet} trained with constrained images, which are inaccurate due to the large domain gap between annotated images and training images. Wang et al.~\cite{wang2017transferring} used the pseudo labels to fine-tune a pre-trained face verification network for AU detection, while Benitez-Quiroz et al.~\cite{benitez2017recognition} introduced a global-local loss to improve robustness on noisy pseudo annotations. Zhao et al.~\cite{zhao2018learning} treated each AU independently during the clustering of re-annotating the pseudo labels but did not take into account the correlations among AUs. All these techniques attempt to work with inaccurate labels and do not exploit accurate AU annotations from other domains like constrained datasets~\cite{zhang2014bp4d,girard2017sayette}, which limits their performance.

Instead of using inaccurate pseudo labels, we consider the approach of transferring AU knowledge from a constrained source domain with accurate AU labels to an unconstrained target domain without AU labels. Recently, self-supervised learning~\cite{wiles2018x2face,koepke2018self-supervised,li2019self-supervised} without requiring annotations is exploited to transform a target image to be a new image with the pose and expression of a source image, in which paired input images with the same identity from a video are required during training. However, a constrained source image and an unconstrained target image with the same identity are unavailable. If training the model using paired same-identity images from the same domain, it will have limited performance of transforming an unconstrained target image driven by a constrained source image, due to the unresolved domain gap.

To make the AU detector trained using source AU labels applicable for the target domain, we can follow the prevailing adversarial domain adaptation approaches. One intuitive way is to learn domain-invariant features~\cite{ganin2016domain,tzeng2017adversarial}. Although this can bring the domains closer, it may result in the loss of AU-related information since AUs are often tangled with poses which can cause the domain shift. Another possible solution is to translate source-domain images to images with target-domain style~\cite{lee2018diverse,zheng2018t2net}. However, only translating the image style fails to reduce other domain shifts caused by pose and occlusion.

\noindent\textbf{Our Solution.} To tackle the above limitations, we propose to map a source image and a target image into a latent domain, which contains the transferred source AU label and the preserved target appearances such as pose, illumination, occlusion, and background. This latent domain is derived by (a) combining source AU-related information with target AU-free information, and (b) learning a mapping that will maximize the performance of target-domain AU detection. Although accurate AU labels are unavailable for the target domain, accurate annotations on highly AU-related landmarks are easily accessible due to contemporary landmark detection methods~\cite{zhu2016face,simon2017hand,wu2018look} with high accuracy comparable to manual labeling. 

We combine the source landmark-related feature with the target landmark-free feature in the latent domain, in which the former contains landmark information and is expected to be AU-related, and the latter discards landmark information and is expected to be AU-free. To alleviate the influence of pose, we choose facial inner-landmarks without contour-landmarks for disentangling landmark-related and landmark-free features. Since there are large domain shifts, it is difficult to simultaneously synthesize realistic images and inherit transferred AU information in the image domain. Instead, we map the unpaired source and target images into a latent feature domain, as illustrated in Fig.~\ref{fig:disentangle}. The latent feature $\tilde{x}^t$ contains source AU-related inner-landmark information and target AU-free global pose and texture information, which is beneficial for training target-domain AU detection.

In particular, 
\textit{the source image is considered to have accurate AU and landmark labels} and \textit{the target image only has accurate landmark labels}. The ``rich'' features learned from images are firstly disentangled into landmark-free features and landmark-related features by a novel landmark adversarial loss, in which the adversarial learning is treated as a multi-player minimax game instead of a two-player minimax game~\cite{goodfellow2014generative}. Then, the landmark-related features of the two images are swapped and combined with the landmark-free features to generate the latent features. A further disentangle-swap-translate process is applied to cross-cyclically reconstruct the original rich features. The entire framework is end-to-end without any post-processing step. During testing, the rich feature of an input target image is simply disentangled, recombined and translated into the latent feature domain for AU detection.

We refer to our framework, AU Detection via Latent Domain, as \textbf{ADLD}. The main contributions of this paper are threefold:
\begin{itemize}
    \item We propose to map the unpaired source and target images into a latent feature domain, which is specialized for the target-domain AU detection. To our knowledge, this is the first work of introducing such an idea for facial AU detection in the wild.
    \item We propose a novel landmark adversarial loss to disentangle the landmark-free feature from the landmark-related feature, in which the adversarial learning for landmark-free feature is treated as a multi-player minimax game.
    \item Extensive experiments demonstrate that our method soundly outperforms lower-bounds and upper-bounds of the basic model, as well as state-of-the-art techniques. The performance of our framework can be further improved by incorporating the pseudo AU labels of the target domain.
\end{itemize}

\section{Related Work}
\label{sec:relatedwork}

We review previous techniques that are most relevant to our work, including facial AU detection in the wild, adversarial domain adaptation, semi-supervised facial AU detection, and feature disentanglement.

\subsection{Facial AU Detection in the Wild}

There are some works exploring the challenging problem of facial AU detection in the wild. Considering accurate annotations of unconstrained images are often unavailable, these methods resort to pseudo AU labels.

On one hand, a pre-trained model for another task can be exploited, since different types of images often have similar characteristics like feature consistency in local regions and approximately Gaussian data distribution. Wang et al.~\cite{wang2017transferring} first pre-trained a face verification network on CASIA-WebFace~\cite{yi2014learning}, then fine-tuned the network on EmotioNet~\cite{benitez2016emotionet} to achieve unconstrained AU detection. Jyoti et al.~\cite{jyoti2019expression} incorporated the features extracted by the network of holistic facial expression recognition into the AU detection network, so as to facilitate AU detection. Ji et al.~\cite{ji2020multiple} fine-tuned two networks pre-trained on face recognition and facial expression recognition datasets respectively, then fused the AU prediction results of two networks.

On the other hand, a few methods focus on improving the robustness on inaccurate annotations. Benitez-Quiroz et al.~\cite{benitez2017recognition} introduced a global-local loss for AU detection with noisy pseudo labels. The local loss aids predicting each AU independently, while the global loss aggregates multiple AUs to probe the co-occurrence among AUs. Zhao et al.~\cite{zhao2018learning} proposed a Weakly Supervised Clustering (WSC) technique to learn an embedding space, which is used to identify visually and semantically similar samples and re-annotate these samples with rank-order clustering. However, each AU is treated independently during clustering, in which the correlations among AUs were ignored. These methods do not explore the use of accurate annotations from other domains, which limits their performance.

\subsection{Adversarial Domain Adaptation}
Adversarial domain adaptation is a prevailing way of transferring knowledge from a source domain to a target domain.

One typical solution is to use an adversarial loss with a domain discriminator to make the features of source and target domains indistinguishable~\cite{tzeng2015simultaneous,ganin2016domain,tzeng2017adversarial,rozantsev2018residual}. Ganin et al.~\cite{ganin2016domain} proposed a Domain-Adversarial Neural Network (DANN) that is shared between domains to learn domain-invariant features. Instead of using a shared network, Tzeng et al.~\cite{tzeng2017adversarial} developed an Adversarial Discriminative Domain Adaptation (ADDA) method by pre-training a network on the source domain and further refining it on the target domain. It minimizes the adversarial loss between the fixed source-domain feature and the trainable target-domain feature. Despite these methods being effective for domain adaptation, enforcing feature domain invariance is infeasible for AU detection. This is because AU-related information may be removed since AUs are often tangled with poses which can cause the domain gap.

Another form involves translating source images into target-style images. For example, Zheng et al.~\cite{zheng2018t2net} presented a method for translating rendered images into the real image domain, with a regularization of identity mapping for real input images. Recently, Wang et al.~\cite{wang2018personalized} utilized a generative adversarial network~\cite{goodfellow2014generative} to synthesize an image with similar appearance to
the target image while retaining AU patterns of the source image, which is a pioneering work of AU detection via adversarial domain adaptation. However, the source and target images processed by this method have similar expressions and are both constrained images, in which only the image differences of AU patterns are considered. Besides, a few gaps between constrained and unconstrained domains like occlusion differences cannot be well resolved by style translation.

\begin{figure*}
\centering\includegraphics[width=\linewidth]{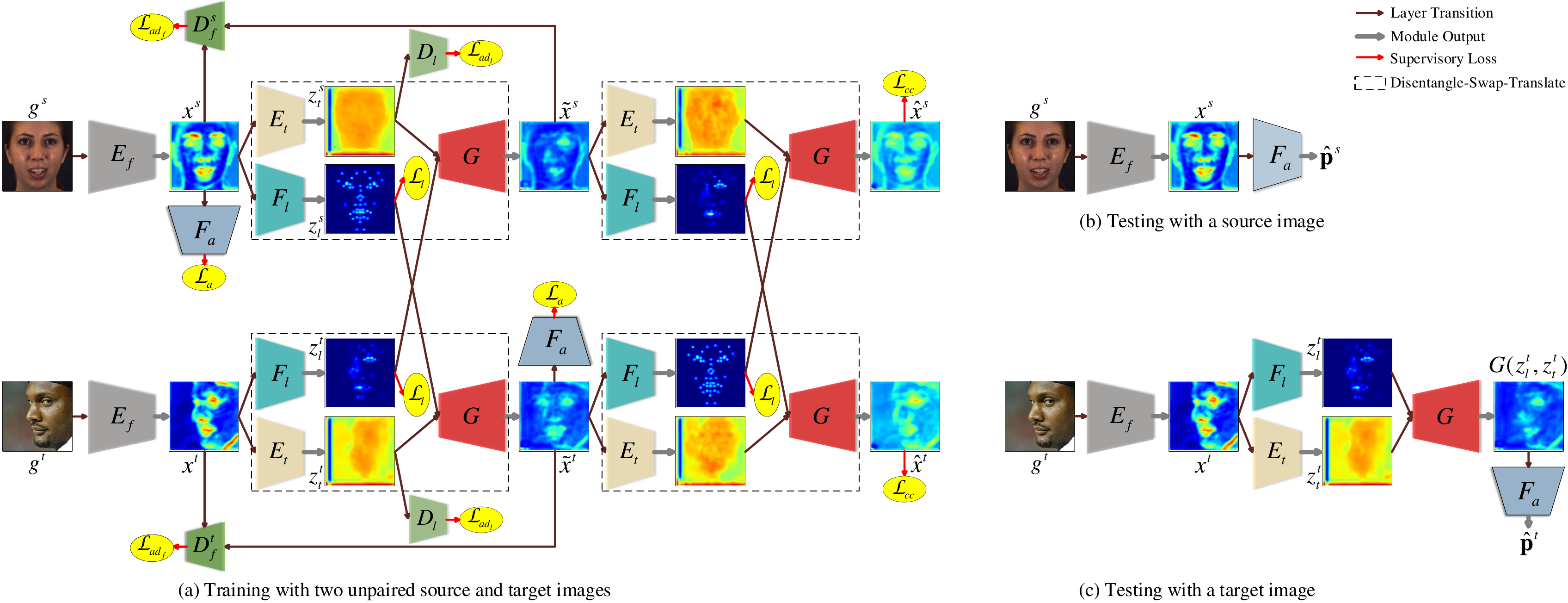}
\caption{The architecture of our ADLD framework, in which $E_f$, $E_t$, $G$, $F_l$, $F_a$ and $D_l$ are shared by source-domain and target-domain input images. \textbf{(a)} During training, given unpaired $g^s$ and $g^t$, $E_f$ first extracts $(x^s, x^t)$ which are further disentangled into $(z^s_{t}, z^t_{t})$ and $(z^s_{l}, z^t_{l})$ by $E_t$ and $F_l$. Then, $G$ combines $z^s_{t}$ and $z^t_{l}$ to generate $\tilde{x}^s$, and combines $z^t_{t}$ and $z^s_{l}$ to generate $\tilde{x}^t$. The disentangle-swap-translate process in the dotted box contains $E_t$, $G$, and $F_l$ with $\mathcal{L}_l$. Another disentangle-swap-translate process is applied to $(\tilde{x}^s, \tilde{x}^t)$ to complete the crossed cycle. \textit{The mapping to the latent feature domain is learned by maximizing the performance of the AU detector $F_a$ given $\tilde{x}^t$.} Note that the self-reconstruction loss $\mathcal{L}_{r}$ is not shown. During testing, we input \textbf{(b)} $x^s$ and \textbf{(c)} $G(z^t_l,z^t_t)$ to $F_a$ for source-domain and target-domain AU detection, respectively.}
\label{fig:ADLD_framework_full}
\end{figure*}

\subsection{Semi-Supervised Facial AU Detection}

Due to the high costs of AU labeling, some AU detection methods use a semi-supervised setting. Specifically, only partial samples have complete AU labels, while remaining samples do not have AU labels or only have labels of partial AUs. Besides, the extreme case is all samples do not have AU labels, in which coarse labels like holistic expression are often used. Since this semi-supervised setting is an alternative way to tackle the lack of AU annotations, we discuss the related works in this section.

One scenario is labels of randomly partial AUs are missing. 
Wu et al.~\cite{wu2015multi} proposed a Multi-Label Learning with Missing Labels (MLML) for AU detection, which assumes the predicted labels to be close for two samples with similar features as well as two classes with similar semantic meanings. 
However, the assumption in MLML is not always correct, as similarity of samples may be due to having the same identity rather than occurring the same AUs. Another scenario is to employ prior knowledge in terms of correlations between AUs and holistic expressions, as well as correlations among AUs. 
To directly aid the learning of AU detector, Zhang et al.~\cite{zhang2018classifier} incorporated prior probabilities including expression-independent and expression-dependent AU probabilities as constraints into the overall objective function. However, applying fixed prior knowledge to all the samples ignores AU dynamics in different samples.

Recently, Niu et al.~\cite{niu2019multi} utilized two networks to generate conditional independent features of different views, and then proposed a multi-label co-regularization loss to enforce the prediction consistency of two views. In this method, a small set of samples with AU labels and a large number of samples without AU labels are from the same domain. Considering each local facial region plays different roles for different AUs and each AU has individual temporal dynamic, Zhang et al.~\cite{zhang2019context} proposed a feature fusion module and a label fusion module by incorporating a learnable task-related context into the attention mechanism. It requires AU intensity labels of peak and valley frames in videos, which is a strict requirement and thus limits its applicability. Different from the above methods, our work is based on domain adaptation and transfers AU knowledge from a constrained domain to an unconstrained domain.


\subsection{Feature Disentanglement}
Feature disentanglement is extensively applied in image or video synthesis, which aims to factorize a feature into different components~\cite{tran2017disentangled,lee2018diverse,shu2018deforming}.

Lee et al.~\cite{lee2018diverse} disentangled representations for image-to-image translation by embedding images into a domain-specific attribute space and a domain-invariant content space that captures shared information across domains. They also employed a cyclic structure~\cite{zhu2017unpaired} to handle unpaired training data. Shu et al.~\cite{shu2018deforming} introduced a generative model to disentangle facial shape and appearance in an unsupervised manner, in which the shape can deform the appearance to generate images. To achieve source-to-target video re-animation, Kim et al.~\cite{kim2018deep} rendered a synthetic target video with the reconstructed head animation parameters from a source video, in which the head animation parameters include disentangled head pose, identity, expression, eye gaze and illumination. In contrast with these methods, our approach proposes a landmark adversarial loss to disentangle the landmark-free feature from the landmark-related feature, and combines the disentangled features in a latent feature domain.

\section{Unconstrained Facial AU Detection}

\subsection{Overview}

Our main goal is to achieve unconstrained facial AU detection, in which the AU occurrence probabilities $\hat{\mathbf{p}}^t$ can be predicted given an unconstrained image $g^t$. The main challenge lies in the training setting that we have access to a collection of constrained images from the source domain with both AU and landmark labels, and also an unpaired collection of unconstrained images from the target domain with only landmark labels. We denote a source image of size $l\times l\times 3$ as $g^s$, with its AU label $\mathbf{p}^s$ and landmark label $\mathbf{q}^s$, while an unpaired target image of the same size is $g^t$ with landmark label $\mathbf{q}^t$. The occurrence probabilities of all $m$ AUs are $\mathbf{p}^s=(p^s_1, \cdots, p^s_m)$, while the x-y positions of all $n$ landmarks are in $\mathbf{q}^s=(q^s_1, q^s_2, \cdots, q^s_{2n-1}, q^s_{2n})$.

Fig.~\ref{fig:ADLD_framework_full} shows the overall architecture of our ADLD framework. During training, our framework consists of two similar paths: top and bottom paths respectively taking in source-domain images and target-domain images. In particular, given two unpaired images $(g^s, g^t)$, we first apply a feature encoder $E_f$ to extract rich features $(x^s, x^t)$. Then we use a texture encoder $E_{t}$ and a landmark detector $F_{l}$ to disentangle the rich features $(x^s, x^t)$ into landmark-free features $(z^s_{t}, z^t_{t})$ and landmark-related features $(z^s_{l}, z^t_{l})$, in which the former are expected to be AU-free and the latter are expected to be AU-related. A generator $G$ is further applied to combine the landmark-free features with the swapped landmark-related features, and translates them to latent features $(\tilde{x}^s, \tilde{x}^t)$. After that, we apply another round of the disentangle-swap-translate process to the latent features to obtain the cross-cyclically reconstructed rich features $(\hat{x}^s, \hat{x}^t)$.

\begin{table}
\centering\caption{Notations in our framework. We only show the source-domain notations, and target-domain notations can be defined similarly.}
\label{tab:notation}
\begin{tabular}{|*{2}{c|}}
\hline
Notation &Definition\\\hline
$g^s$ &source-domain input image\\\hline
$l$ &width of input image\\\hline
$p^s_j$ &\tabincell{c}{ground-truth occurrence probability\\of the $j$-th source-domain AU}\\\hline
$\hat{p}^s_j$ &\tabincell{c}{predicted occurrence probability\\of the $j$-th source-domain AU}\\\hline
$q^s_{2i-1}, q^s_{2i}$ &\tabincell{c}{ground-truth x- and y-coordinates\\of the $i$-th source-domain landmark}\\\hline
$y^s_i$ &\tabincell{c}{ground-truth 1-D location index\\of the $i$-th source-domain landmark}\\\hline
$d$ &width of landmark response map\\\hline
$m$ &number of AUs\\\hline
$n$ &number of landmarks\\\hline
$x^s$ &source-domain rich feature\\\hline
$z^s_{t}$ &landmark-free feature from $x^s$\\\hline
$z^s_l$ &landmark-related feature from $x^s$\\\hline
$\tilde{x}^s$ &source-domain latent feature\\\hline
$\hat{x}^s$ &cross-cyclically reconstructed rich feature for $x^s$\\\hline
\end{tabular}
\end{table}

The key to the AU label transfer from source to target images lies in the combination of the target landmark-free feature $z^t_{t}$, which contains the target global pose and texture for
adapting to unconstrained conditions of the target domain, with the source landmark-related feature $z^s_{l}$, which brings over the associated source AU label. In this way, we use the transferred AU labels to train an AU detector which can adapt to the unconstrained target domain. By maximizing the performance of the AU detector $F_a$ given $\tilde{x}^t$, we can learn the mapping from source and target domains to the latent feature domain. The landmark discriminator $D_{l}$ is used to ensure the landmark-free feature cannot predict the locations of landmarks so as to be disentangled from the landmark-related feature. The feature discriminators $\{D^s_{f}, D^t_{f}\}$ aim to discriminate between the rich features $(x^s, x^t)$ and the latent features $(\tilde{x}^s, \tilde{x}^t)$ in order to bring them closer. We denote the domains of features and labels using the corresponding capitals, e.g., domain $X^T$ for $x^t$. The main notations are summarized in Table~\ref{tab:notation}.

\subsection{AU Label Transfer}

\subsubsection{Definition of AU-Related Landmarks}

\begin{figure}
\centering\includegraphics[width=\linewidth]{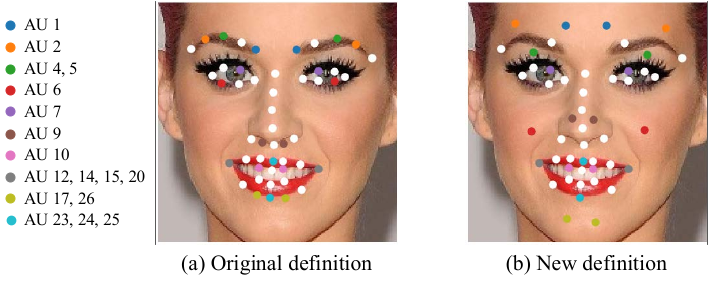}
\caption{Definition of AU-related landmarks, in which two landmarks in the same color correspond to the two centers of a certain AU. We replace these landmarks in the \textbf{(a)} original definition with their predefined AU centers in the \textbf{(b)} new definition.}
\label{fig:landmark_definition}
\end{figure}

A few previous works~\cite{li2018eac,shao2018deep} exploit facial landmarks to predefine the locations of AU centers based on prior knowledge, as defined in Table~\ref{tab:au_definition}. Some AU centers are exactly on the locations of landmarks, and other AU centers have certain offsets from the locations of landmarks. The corresponding landmarks of these predefined AU centers are from $49$ facial inner-landmarks~\cite{xiong2013supervised}, as illustrated in Fig.~\ref{fig:landmark_definition}(a). Considering the predefined AU centers can be used to extract highly AU-related features so as to facilitate AU detection, we use these AU centers to replace their corresponding landmarks, as shown in Fig.~\ref{fig:landmark_definition}(b).

\begin{table}
\centering\caption{Rules for defining the locations of AU centers, which are applicable to an aligned face image with eye centers on the same horizontal line. ``Scale'' denotes the distance between the inner corners of eyes.}
\label{tab:au_definition}
\begin{tabular}{|*{3}{c|}}
\hline
AU &Description &Location\\\hline
1 &Inner brow raiser &1/2 scale above inner brow\\\hline
2 &Outer brow raiser &1/3 scale above outer brow\\\hline
4 &Brow lowerer &1/3 scale below brow center\\\hline
5 &Upper lid raiser &1/3 scale below brow center\\\hline
6 &Cheek raiser &1 scale below eye bottom\\\hline
7 &Lid tightener &Eye\\\hline
9 &Nose wrinkler &1/2 scale above nose bottom\\\hline
10 &Upper lip raiser &Upper lip center\\\hline
12 &Lip corner puller &Lip corner\\\hline
14 &Dimpler &Lip corner\\\hline
15 &Lip corner depressor &Lip corner\\\hline
17 &Chin raiser &1/2 scale below lip\\\hline
20 &Lip stretcher &Lip corner\\\hline
23 &Lip tightener &Lip center\\\hline
24 &Lip pressor &Lip center\\\hline
25 &Lips part &Lip center\\\hline
26 &Jaw drop &1/2 scale below lip\\\hline
\end{tabular}
\end{table}

Since the correlations among different facial regions are beneficial for AU detection~\cite{shao2019facial}, other landmarks are also employed. Note that these $49$ landmarks do not contain facial contour-landmarks which are on the facial global contour. In this way, the learned landmark-free feature can discard AU-related information in facial inner regions while preserving AU-free facial global pose. Besides, the new landmark definition in Fig.~\ref{fig:landmark_definition}(b) is applied for all the different datasets, even if some AUs in Table~\ref{tab:au_definition} are not evaluated due to the lack of their annotations. This is because the detection of a certain AU can benefit from the correlations with other AUs.

\subsubsection{Disentanglement of Landmark-Free and Landmark-Related Features}

Taking $x^t$ as an example, we want it to be disentangled into the landmark-free feature $z^t_{t}$ and the landmark-related feature $z^t_{l}$, in which the former is free of facial inner-landmark information and the latter contains inner-landmark information.

\noindent\textbf{Landmark-Free Feature.}
To remove inner-landmark information for the landmark-free feature, we introduce the landmark discriminator $D_l$ as the adversary of the texture encoder $E_{t}$. Since adversarial learning~\cite{goodfellow2014generative} for cross entropy loss is widely used in feature disentanglement~\cite{tran2017disentangled,shu2018deforming}, we regard facial landmark detection as a classification problem~\cite{honari2016recombinator,xiao2016robust} instead of a regression problem~\cite{zhang2015learning,shao2019deep}. Specifically, the output of $D_l$ is $n$ feature maps, each of which can be seen as a response map with a size of $d\times d\times 1$ for each landmark. Each position in the response map is considered as one class and the total number of classes is $d^2$. The class label of the $i$-th landmark is defined as
\begin{equation} \label{eq:ysi}
    y^t_i= (\lfloor q^t_{2i}d/l\rceil-1)d+\lfloor q^t_{2i-1}d/l\rceil,
\end{equation}
where $\lfloor \cdot \rceil$ denotes the operation of rounding a number to the nearest integer, and $i=1, \cdots, n$. Eq.~\eqref{eq:ysi} is used for converting the landmark detection from a regression problem to a classification problem, in which the ground-truth x- and y-coordinates of a landmark at $l\times l$ scale are transformed to a 1-D location index at $1\times d^2$ scale.

Similar to the conventional adversarial loss~\cite{goodfellow2014generative} with the form of binary cross entropy loss, we define the landmark adversarial loss as a multi-class cross entropy loss of the multi-player minimax game:
\begin{equation}\label{eq:advl}
\begin{aligned}
    \mathcal{L}_{{ad}_l}&(E_{t},D_l,X^T,\mathbf{Y}^T)=\\
    &\mathbb{E}_{x^t\sim X^T}[\frac{1}{n}\sum_{i=1}^n\sum_{k=1}^{d^2}\mathbbm{1}_{[k=y^t_i]}\log(\sigma(D_{l}^{(i,k)}(E_{t}(x^t))))],
\end{aligned}
\end{equation}
where $E_{t}(x^t)=z^t_{t}$, $D_{l}^{(i,k)}(\cdot)$ is the $k$-th value in the $i$-th response map output by $D_{l}$, $\mathbbm{1}_{[\cdot]}$ denotes the indicator function, and $\sigma(\cdot)$ denotes the softmax function that is applied across spatial locations for each response map. However, the two-player minimax game~\cite{goodfellow2014generative} designed for binary cross entropy loss does not work for this multi-class cross entropy loss.

We propose a novel strategy to solve this multi-player minimax game in adversarial learning. While keeping the same adversarial principle, we train $D_l$ by minimizing:
\begin{equation}\label{eq:EDl}
\begin{aligned}
    &\mathbb{E}_{x^t\sim X^T}[\frac{1}{nd^2}\sum_{i=1}^n\sum_{k=1}^{d^2}(\mathbbm{1}_{[k\neq y^t_i]}\lVert D_{l}^{(i,k)}(E_{t}(x^t))\rVert^2_2+\\
    &\qquad\qquad\qquad\quad\ \ \mathbbm{1}_{[k=y^t_i]}\lVert D_{l}^{(i,k)}(E_{t}(x^t))-1\rVert^2_2)],
\end{aligned}
\end{equation}
where we encourage $D_{l}$ to generate $1$ at the ground-truth landmark locations while generating $0$ at the other locations. Conversely, we train $E_{t}$ by minimizing:
\begin{equation}\label{eq:EEt}
    \mathbb{E}_{x^t\sim X^T}[\frac{1}{nd^2}\sum_{i=1}^n\sum_{k=1}^{d^2}\lVert D_{l}^{(i,k)}(E_{t}(x^t))-\frac{1}{d^2}\rVert^2_2],
\end{equation}
where $E_{t}$ tries to remove the landmark information as much as possible so that $D_{l}$ will generate the same probability $1/d^2$ for all possible landmark locations.

Such least-squares loss in Eqs.~\eqref{eq:EDl} and \eqref{eq:EEt} is often used in adversarial learning due to its stability~\cite{mao2017least}. The combination of Eq.~\eqref{eq:EDl} and Eq.~\eqref{eq:EEt} completely defines the landmark adversarial loss $\mathcal{L}_{{ad}_l}(E_{t},D_l,X^T,\mathbf{Y}^T)$. In Fig.~\ref{fig:ADLD_framework_full}(a), we can observe that $z^t_{t}$ contains AU-free information including global pose and texture, which are beneficial for the latent feature $\tilde{x}^t$ to adapt to unconstrained conditions of the target domain. Besides, the gradients from $E_t$ are set to not be back-propagated to $E_f$ and $G$ for avoiding the adversarial training between $E_t$ and $D_l$ impacts the learning of $x^t$ and $\tilde{x}^t$, respectively.

\noindent\textbf{Landmark-Related Feature.}
To extract the landmark-related feature, we employ the landmark detector $F_l$ to predict the locations of facial inner-landmarks. By treating the landmark detection as a classification problem, we define the landmark classification loss as
\begin{equation} \label{eq:Ll}
\begin{aligned}
    \mathcal{L}_{l}&(F_{l},X^T,\mathbf{Y}^T)=\\
    &-\mathbb{E}_{x^t\sim X^T}[\frac{1}{n}\sum_{i=1}^n\sum_{k=1}^{d^2}\mathbbm{1}_{[k=y^t_i]}\log(\sigma(F_{l}^{(i,k)}(x^t)))],
\end{aligned}
\end{equation}
where $F_l$ also outputs $n$ response maps similar to $D_l$. Minimizing Eq.~\eqref{eq:Ll} encourages the $i$-th response map to have the highest response $\sigma(F_{l}^{(i,y^t_i)}(x^t))$ at the location $(\lfloor q^t_{2i-1}d/l\rceil, \lfloor q^t_{2i}d/l\rceil)$ while having near-zero responses at other locations.

To make the landmark-related feature $z^t_{l}$ contain facial inner shape information, we sum the response maps of all $n$ landmarks element-wise:
\begin{equation}
\label{eq:shape}
    z^t_{l}= \oplus_{i=1}^n \sigma(F_{l}^{(i)}(x^t)),
\end{equation}
where $\oplus$ denotes element-wise sum. We express Eq.~\eqref{eq:shape} with a simplified form $z^t_{l}=\check{F}_l(x^t)$. The landmark-related feature is enforced to only have high responses at the landmark locations while discarding other AU-free information, as shown in Fig.~\ref{fig:ADLD_framework_full}(a).

\subsubsection{AU Detection in Latent Feature Domain}

Since the landmark-related feature contains AU-related information, we can inherit the source AU label by introducing the source landmark-related feature $z^s_l$. In particular, we swap the landmark-related features $z^s_l$ and $z^t_l$, and input $z^s_l$ and $z^t_{t}$ to the generator $G$ to generate the latent feature $\tilde{x}^t$:
\begin{equation}
    \tilde{x}^t=G(\check{F}_l(x^s), E_t(x^t)),
\end{equation}
where the channels of $z^s_l$ and $z^t_{t}$ are concatenated to input to $G$. $\tilde{x}^t$ in the latent feature domain is expected to include preserved AU-free information from $x^t$, and transferred AU-related information with AU and landmark labels from $x^s$. To enforce $\tilde{x}^t$ to inherit source AU-related information, we apply $\mathcal{L}_{l}(F_{l},\tilde{X}^T,\mathbf{Y}^S)$. In Fig.~\ref{fig:ADLD_framework_full}(a), at each training iteration, the parameters of $F_l$ are updated for $x^s$ and $x^t$, while fixed for $\tilde{x}^s$ and $\tilde{x}^t$ so that $F_l$ only used for constraining their generation. This is to avoid that $F_l$ is influenced by the generation of latent features, which will weaken the effect of constraint.

Then, we achieve the target-domain AU detection by applying the AU detector $F_a$ on $\tilde{x}^t$ with an AU detection loss:
\begin{equation}
\label{eq:au_loss_t}
\begin{aligned}
    \mathcal{L}_{a}&(F_a, \tilde{X}^T, \mathbf{P}^S)=\\
    &-\mathbb{E}_{\tilde{x}^t\sim \tilde{X}^T}[\sum_{j=1}^m w^s_j(p^s_j\log\hat{p}^s_j+(1-p^s_j)\log(1-\hat{p}^s_j))],
\end{aligned}
\end{equation}
where $p^s_j$ is ground-truth occurrence probability of the $j$-th AU transferred from $x^s$, $\hat{p}^s_j=\delta({F_a}^{(j)}(\tilde{x}^t))$ is predicted occurrence probability of the $j$-th AU, $\delta(\cdot)$ is the sigmoid function, and $w^s_j$ is a weight parameter~\cite{shao2018deep} for alleviating the data imbalance problem. We choose $w^s_j = (1/r^s_j)/\sum_{u=1}^{m}(1/r^s_u)$, where $r^s_j$ is the occurrence rate of the $j$-th AU in source-domain training set. With Eq.~\eqref{eq:au_loss_t}, we learn the mapping from source and target domains to the latent feature domain by maximizing the performance of target-domain AU detection. Although we do not focus on source-domain AU detection, $\tilde{x}^s$ is also obtained in the latent feature domain due to the symmetric structure in our ADLD framework, as shown in Fig.~\ref{fig:ADLD_framework_full}(a).

\subsubsection{Reliability Constraints on Latent Feature Domain}

To obtain a reliable latent feature domain, we want the latent feature domain has a similar structure to the source domain and the target domain. To encourage the latent features to be indistinguishable from the rich features, we impose two feature discriminators $D^s_f$ and $D^t_f$ with a feature adversarial loss $\mathcal{L}_{{ad}_f}$ for source and target domains, respectively. $\mathcal{L}_{{ad}_f}$ for $\tilde{x}^t$ is defined as
\begin{equation} \label{eq:advf}
\begin{aligned}
    \mathcal{L}_{{ad}_f}&(F_l,E_t,G,D^t_f,X^T)=\\
    &\mathbb{E}_{x^t\sim X^T}[\log D^t_f(x^t)]+\mathbb{E}_{\tilde{x}^t\sim \tilde{X}^T}[\log(1- D^t_f(\tilde{x}^t))].
\end{aligned}
\end{equation}
For stable adversarial learning, in our implementation we use the least-squares loss~\cite{mao2017least} to train $\mathcal{L}_{{ad}_f}$. Particularly, we train $D^t_f$ by minimizing $\mathbb{E}_{x^t\sim X^T}[\lVert D^t_f(x^t)-1\lVert^2_2]+\mathbb{E}_{\tilde{x}^t\sim \tilde{X}^T}[\lVert D^t_f(\tilde{x}^t)\lVert^2_2]$, and train $G$ by minimizing $\mathbb{E}_{\tilde{x}^t\sim \tilde{X}^T}[\lVert D^t_f(\tilde{x}^t)-1\lVert^2_2]$.

As illustrated in Fig.~\ref{fig:ADLD_framework_full}(c), the
rich feature $x^t$ of an input target image $g^t$ is disentangled, recombined and translated to be a self-reconstructed latent feature $G(z^t_l,z^t_t)$ during testing. Similarly, we expect this self-reconstructed latent feature to be similar to the rich feature during training by using a self-reconstruction loss:
\begin{equation} \label{eq:Lr}
    \mathcal{L}_{r}(F_l,E_t,G,X^T)=\mathbb{E}_{x^t\sim X^T} [\lVert G(z^t_l,z^t_t)-x^t\rVert_1].
\end{equation}
Besides, considering the effectiveness of cyclic structure~\cite{zhu2017unpaired,lee2018diverse} for unpaired training data, we employ a cross-cycle consistency loss $\mathcal{L}_{cc}$ to encourage the cross-cyclically reconstructed rich feature to be similar to the rich feature:
\begin{equation} \label{eq:Lcc}
    \mathcal{L}_{cc}(F_l,E_{t},G,X^T,X^S)=\mathbb{E}_{x^t\sim X^T,x^s\sim X^S} [\lVert \hat{x}^t-x^t\rVert_1],
\end{equation}
where $\hat{x}^t = G(\check{F}_l(\tilde{x}^s),E_t(\tilde{x}^t))$. With $\mathcal{L}_{{ad}_f}$, $\mathcal{L}_{r}$ and $\mathcal{L}_{cc}$, we can generate a reliable latent feature domain specialized for target-domain AU detection.

\subsection{Overall Objective Function}

As shown in Fig.~\ref{fig:ADLD_framework_full}(a), the losses introduced above are applied for both source and target images in our ADLD framework. Specifically, $\mathcal{L}_{l}(F_l, X^S, \mathbf{Y}^S)$ and $\mathcal{L}_{l}(F_l, X^T, \mathbf{Y}^T)$ are used for training the landmark detector $F_l$, and $\mathcal{L}_{l}(F_{l},\tilde{X}^S,\mathbf{Y}^T)$ and $\mathcal{L}_{l}(F_{l},\tilde{X}^T,\mathbf{Y}^S)$ are only used for constraining the generation of latent features $\tilde{x}^s$ and $\tilde{x}^t$. $\mathcal{L}_{a}(F_a, X^S, \mathbf{P}^S)$ and $\mathcal{L}_{a}(F_a, \tilde{X}^T, \mathbf{P}^S)$ are used for training the AU detector $F_a$. The remaining losses defined in Eqs.~\eqref{eq:advl},~\eqref{eq:advf},~\eqref{eq:Lr},~\eqref{eq:Lcc} are also applied to the source image: $\mathcal{L}_{{ad}_l}(E_{t},D_l,X^S,\mathbf{Y}^S)$, $\mathcal{L}_{{ad}_f}(F_l,E_t,G,D^s_f,X^S)$, $\mathcal{L}_{r}(F_l,E_t,G,X^S)$, $\mathcal{L}_{cc}(F_l,E_{t},G,X^S,X^T)$.

Combining all the losses, we yield the overall objective function:
\begin{equation}
\label{eq:full}
\begin{aligned}
    &\min_{\{F_{a},F_l\}}\min_{\{E_f,E_t,G\}}\max_{\{D_l,D^s_{f},D^t_{f}\}}\mathcal{L}_{ADLD} = \\ &\mathcal{L}_{a}+\lambda_l\mathcal{L}_{l}+\lambda_{{ad}_l}\mathcal{L}_{{ad}_l}+\lambda_{{ad}_f}\mathcal{L}_{{ad}_f}+\lambda_{r}\mathcal{L}_{r}+\lambda_{cc}\mathcal{L}_{cc},
\end{aligned}
\end{equation}
where the hyper-parameters $\lambda_{(.)}$ control the importance of each loss term. Our framework is trainable end-to-end, in which all the network modules are trained jointly. At test time, the inputs of $F_{a}$ are source rich feature $x^s$ and target self-reconstructed latent feature $G(z^t_l,z^t_t)$ for given source and target images, respectively. This inference process is consistent with the training process, which is beneficial for AU detection in both source and target domains.

\begin{table*}
\centering\caption{AU occurrence rates ($\%$) in the training sets of source domain. ``-'' denotes the dataset does not contain this AU. The AUs with occurrence rates larger than $6\%$ are shown in bold.
}
\label{tab:au_occurrence}
\begin{tabular}{|*{12}{c|}}
\hline
AU &1 &2 &4 &5 &6 &9 &12 &17 &20 &25 &26\\
\hline
BP4D~\cite{zhang2014bp4d} &\textbf{18.4} &\textbf{14.6} &\textbf{19.8} &3.3 &\textbf{44.0} &5.7 &\textbf{54.0} &\textbf{34.2} &2.6 &- &-\\
\hline
GFT~\cite{girard2017sayette} &4.1 &\textbf{14.7} &4.1 &2.5 &\textbf{29.2} &1.5 &\textbf{30.3} &\textbf{28.7} &- &- &-\\
\hline
DISFA~\cite{mavadati2013disfa} &4.3 &3.6 &\textbf{12.2} &0.8 &\textbf{7.2} &3.1 &\textbf{12.9} &4.4 &2.0 &\textbf{26.2} &\textbf{8.8}\\
\hline
Pain~\cite{lucey2011painful} &- &- &2.4 &- &\textbf{8.7} &0.8 &\textbf{10.9} &- &1.0 &4.4 &4.4\\
\hline
\end{tabular}
\end{table*}

\section{Experiments}

\subsection{Datasets and Settings}
\subsubsection{Datasets}

In our experiments, we utilized four popular AU detection datasets BP4D~\cite{zhang2014bp4d}, GFT~\cite{girard2017sayette}, DISFA~\cite{mavadati2013disfa} and UNBC-McMaster Shoulder Pain~\cite{lucey2011painful} for the constrained source domain, and utilized challenging EmotioNet~\cite{benitez2016emotionet} and VGGFace2~\cite{cao2018vggface2} for the unconstrained target domain, respectively. Note that we evaluate frame-level AU detection, and thus other datasets with only video-level annotations like CK+~\cite{lucey2010extended} are not used.

\begin{itemize}
\item \textbf{BP4D} comprises of $328$ videos with $41$ subjects, each of whom participates in $8$ sessions. These videos contain both AU and landmark annotations, which were captured in constrained conditions with near-frontal faces in good illumination and simple backgrounds. We removed a few frames without AU and landmark annotations, and partitioned the remaining frames into a training set with $100,767$ images of $28$ subjects, a validation set with $24,869$ images of $7$ subjects and a test set with $20,940$ images of $6$ subjects.

\item \textbf{GFT} includes $96$ subjects in $32$ three-subject groups with unscripted social interactions, in which each subject was captured using a video with both AU and landmark annotations. Although the captured frames show moderate out-of-plane poses, they are still in constrained conditions with good illumination and simple backgrounds. There are a few frames without AU annotations. We ignored these frames, and partitioned the remaining frames into a training set with $83,346$ images of $60$ subjects, a validation set with $24,145$ images of $18$ subjects and a test set with $24,621$ images of $18$ subjects.

\item \textbf{DISFA} consists of $27$ subjects, each of whom was recorded by one video. Each frame was labeled with $66$ facial landmarks, which includes the $49$ landmarks in Fig.~\ref{fig:landmark_definition}(a), as well as AU intensities on a six-point ordinal scale from $0$ to $5$. Following the setting in~\cite{li2018eac,shao2018deep}, we treated AU intensities equal or greater than $2$ as occurrence, while treated others as non-occurrence. The frames were partitioned into a training set with $82,971$ images of $18$ subjects, a validation set with $19,275$ images of $4$ subjects and a test set with $23,898$ images of $5$ subjects.

\item \textbf{UNBC-McMaster Shoulder Pain} was captured with $200$ videos from $25$ subjects suffering from shoulder pain. Each frame was annotated with $66$ landmarks as well as AU intensities ranging from $0$ to $5$. Similar to the setting in DISFA, AU intensities equal or greater than $2$ were considered as occurrence, while others were considered as non-occurrence. The frames were partitioned into a training set with $34,025$ images of $18$ subjects, a validation set with $6,269$ images of $3$ subjects and a test set with $8,104$ images of $4$ subjects. We denote this dataset as \textbf{Pain} in the following sections.

\item \textbf{EmotioNet} contains about one million training and validation images collected from the Internet, and exhibits unconstrained variations of expression, pose, illumination and occlusion. The AU labels of training images were automatically annotated by~\cite{benitez2016emotionet} and those of validation images were manually annotated by certified experts. Since landmark annotations were not provided, we employed a powerful landmark detection library OpenPose~\cite{simon2017hand} to annotate $49$ facial landmarks as defined in Fig.~\ref{fig:landmark_definition}(a) for each image, in which the images failed to be detected with landmarks were removed. We randomly selected $100,767$ training images as a training set, and split the validation images into a validation set with $10,544$ images and a test set with $ 10,544$ images. Note that the training set has inaccurate pseudo AU labels, while the validation set and the test set have accurate manual AU labels.

\item \textbf{VGGFace2} is a large-scale face recognition dataset, which consists of $3.31$ million images with large variations in pose, age and illumination. We also use OpenPose~\cite{simon2017hand} to annotate $49$ facial landmarks for each image. Since VGGFace2 does not have manual AU labels as well as pseudo AU labels, it is applied in the scenario of only using landmark labels for the target domain. We randomly selected $100,767$ images as a training set, and use the validation and test sets of EmotioNet for validation and testing, respectively.
\end{itemize}

\subsubsection{Evaluation Metrics}

The common AUs of BP4D and EmotioNet are AUs 1, 2, 4, 5, 6, 9, 12, 17 and 20, the common AUs of GFT and EmotioNet are AUs 1, 2, 4, 5, 6, 9, 12 and 17, the common AUs of DISFA and EmotioNet are AUs 1, 2, 4, 5, 6, 9, 12, 17, 20, 25, 26, and the common AUs of Pain and EmotioNet are AUs 4, 6, 9, 12, 20, 25, 26.

The AU occurrence rates in the training sets of source domain are shown in Table~\ref{tab:au_occurrence}. We can see that some AUs like AU 5 and AU 20 have very low occurrence rates, while other AUs like AU 6 and AU 12 have high occurrence rates. Similar to~\cite{zhao2018learning}, to alleviate this data imbalance issue, we chose the AUs with occurrence rates in the source-domain training set larger than $6\%$ to evaluate our framework. In this way, we used AUs 1, 2, 4, 6, 12 and 17 for BP4D, used AUs 2, 6, 12 and 17 for GFT, used AUs 4, 6, 12, 25, 26 for DISFA, and used AUs 6 and 12 for Pain.

Following the previous techniques~\cite{li2018eac,shao2018deep}, we report the frame-based F1-score (F1-frame) for AU detection; meanwhile the average result over all AUs (abbreviated as Avg) is also presented. In the following sections, the F1-frame results are reported in percentages with ``$\%$'' omitted.

\subsubsection{Implementation Details}

Our ADLD framework consists of $F_a$, $F_l$, $E_f$, $E_t$, $G$, $D_l$, $D^s_f$ and $D^t_f$. Specifically, $F_a$ uses an independent branch to estimate the occurrence probability of each AU, in which each branch contains $4$ convolutional layers followed by a global average pooling layer~\cite{lin2013network} and a one-dimensional fully-connected layer. $F_l$ and $D_l$ have the same structure with $5$ convolutional layers, where the last layer has $n$ channels. Other modules are mainly composed of convolutional layers, in which $D^s_f$ and $D^t_f$ share the same structure. For $E_f$ and $F_a$ which are related to the AU detection task, each convolutional layer is followed by Batch Normalization~\cite{ioffe2015batch} and Parametric Rectified Linear Unit (PReLU)~\cite{he2015delving}. For $F_l$, $E_t$, $G$, $D_l$, $D^s_f$ and $D^t_f$ with generation and discrimination, each convolutional layer is followed by Instance Normalization~\cite{ulyanov2016instance} and PReLU. To facilitate feature translation, the Tanh function is applied to the outputs of $E_f$, $E_{t}$ and $G$.

Our framework was trained using PyTorch~\cite{paszke2019pytorch}. Similar to Shao et al.~\cite{shao2018deep}, each sample image was aligned to $200\times 200\times 3$ using similarity transformation and further randomly cropped into $l\times l\times 3$ and mirrored. In our experiments, the number of landmarks $n$, the crop size $l$ and the width of landmark response map $d$ are set to $49$, $176$ and $44$, respectively. The numbers of channels for $x^t$, $z^t_t$ and $z^t_l$ are $64$, $64$ and $1$, respectively. The hyper-parameters of different loss terms are set via obtaining overall best performance on validation sets: $\lambda_l=0.6$, $\lambda_{{ad}_l}=400$, $\lambda_{{ad}_f}=1.2$, $\lambda_{r}=3$ and $\lambda_{cc}=40$. We used the Adam solver~\cite{kingma2014adam}, setting $\beta_1=0.5$, $\beta_2=0.9$ and an initial learning rate of $5\times10^{-5}$ for $E_{t}$, $G$, $D_l$, $D^s_f$ and $D^t_f$, as well as $\beta_1=0.95$, $\beta_2=0.999$ and an initial learning rate of $10^{-4}$ for $E_f$, $F_a$ and $F_l$. The learning rates were unchanged during the first $5$ epochs and linearly decayed during the next $5$ epochs. More details can be found in our code \textit{https://github.com/ZhiwenShao/ADLD}.

\begin{table*}
\centering\caption{F1-frame results of lower-bounds and upper-bounds of the basic model, as well as our approach on the test sets of BP4D~\cite{zhang2014bp4d} and EmotioNet~\cite{benitez2016emotionet}. The best results are shown in bold.}
\label{tab:comp_baseline}
\begin{tabular}{|*{15}{c|}}
\hline
\multirow{2}*{AU} &\multicolumn{7}{c|}{BP4D (source domain)} &\multicolumn{7}{c|}{EmotioNet (target domain)}\\
\cline{2-15}&1 &2 &4 &6 &12 &17 &\textbf{Avg} &1 &2 &4 &6 &12 &17 &\textbf{Avg}\\\hline
LI$_{s(a,l)}$ &57.8 &24.7 &\textbf{67.2} &75.2 &\textbf{84.6} &60.9 &61.7 &12.0 &6.9 &11.4 &27.9 &53.5 &3.5 &19.2\\
LI$^{t(l)}_{s(a,l)}$ &\textbf{65.9} &\textbf{39.5} &59.8 &\textbf{78.4} &75.7 &62.4 &\textbf{63.6} &19.0 &8.7 &21.5 &38.1 &58.4 &7.3 &25.5\\
\textbf{ADLD} &50.5 &35.7 &61.8 &74.1 &75.2 &\textbf{69.0} &61.0 &\textbf{19.8} &\textbf{25.2} &\textbf{31.0} &\textbf{58.2} &\textbf{78.3} &\textbf{8.6} &\textbf{36.8}\\\hline
UI$^{t(a,l)}$ &5.1 &2.8 &35.9 &73.5 &81.3 &0.7 &33.2 &14.7 &11.4 &41.5 &49.4 &75.8 &11.4 &34.0\\
UI$^{t(a,l)}_{s(a,l)}$ &24.1 &31.3 &62.1 &77.4 &79.9 &39.8 &52.4 &15.3 &9.1 &38.5 &48.9 &74.9 &4.4 &31.9\\
\textbf{ADLD-Full} &\textbf{45.7} &\textbf{37.3} &\textbf{63.6} &\textbf{81.7} &\textbf{82.8} &\textbf{64.6} &\textbf{62.6} &\textbf{30.7} &\textbf{26.1} &\textbf{48.1} &\textbf{60.7} &\textbf{77.6} &\textbf{11.5} &\textbf{42.4}\\
\hline
\end{tabular}
\end{table*}

\begin{table*}
\centering\caption{F1-frame results of lower-bounds and upper-bounds of the basic model, as well as our approach on the test sets of GFT~\cite{girard2017sayette} and EmotioNet~\cite{benitez2016emotionet}.}
\label{tab:comp_baseline_gft}
\begin{tabular}{|*{11}{c|}}
\hline
\multirow{2}*{AU} &\multicolumn{5}{c|}{GFT (source domain)} &\multicolumn{5}{c|}{EmotioNet (target domain)}\\
\cline{2-11}&2 &6 &12 &17 &\textbf{Avg} &2 &6 &12 &17 &\textbf{Avg}\\\hline
LI$_{s(a,l)}$ &44.7 &70.7 &\textbf{83.7} &53.6 &63.2 &6.0 &32.7 &58.0 &5.2 &25.5\\
LI$^{t(l)}_{s(a,l)}$ &\textbf{47.2} &76.8 &79.5 &\textbf{58.2} &\textbf{65.4} &8.6 &44.2 &69.2 &6.4 &32.1\\
\textbf{ADLD} &39.8 &\textbf{79.3} &81.4 &54.9 &63.8 &\textbf{17.4} &\textbf{59.3} &\textbf{80.2} &\textbf{9.5} &\textbf{41.6}\\\hline
UI$^{t(a,l)}$ &1.0 &69.9 &73.4 &9.0 &38.3 &14.8 &53.6 &80.4 &11.0 &39.9\\
UI$^{t(a,l)}_{s(a,l)}$ &27.1 &71.5 &82.2 &47.4 &57.0 &18.7 &51.7 &80.5 &10.7 &40.4\\
\textbf{ADLD-Full} &\textbf{43.9} &\textbf{73.3} &\textbf{83.2} &\textbf{57.1} &\textbf{64.4} &\textbf{21.9} &\textbf{64.9} &\textbf{85.4} &\textbf{11.6} &\textbf{46.0}\\\hline
\end{tabular}
\end{table*}

\subsection{Our Framework vs. Lower-Bounds and Upper-Bounds of the Basic Model}

In our training setting, we made use of the source images with both AU and landmark labels, and the target images with only landmark labels. We treat a network composed of the AU detection related modules from our framework as the \textbf{Basic Model}. Specifically, it comprises $E_f$ followed by two parallel modules $F_a$ and $F_l$, in which the lower-bounds use the same training setting and the upper-bounds further use target-domain pseudo AU labels. To validate our framework, we expect that our method performs better than both lower-bounds and upper-bounds of the basic model for target-domain AU detection.

In particular, there are two lower-bounds of the basic model: \textbf{LI}$_{s(a,l)}$ and \textbf{LI}$^{t(l)}_{s(a,l)}$, in which the former was trained with $\mathcal{L}_{a}(F_a, X^S, \mathbf{P}^S)$ and $\mathcal{L}_{l}(F_{l},X^S,\mathbf{Y}^S)$ using only the source images with AU and landmark labels, and the latter further utilizes the target images with landmark labels by adding $\mathcal{L}_{l}(F_{l},X^T,\mathbf{Y}^T)$. By using pseudo AU labels of the target images, there are two upper-bounds of the basic model: \textbf{UI}$^{t(a,l)}$ and \textbf{UI}$^{t(a,l)}_{s(a,l)}$, in which the former only employs the target images with pseudo AU labels and landmark labels, and the latter employs images with all available labels of both domains. Moreover, our ADLD framework in Fig.~\ref{fig:ADLD_framework_full}(a) can be naturally extended to the scenario with target-domain pseudo AU labels by applying $\mathcal{L}_{a}(F_a, X^T, \mathbf{P}^T)$ and $\mathcal{L}_{a}(F_a, \tilde{X}^S, \mathbf{P}^T)$, which is denoted as \textbf{ADLD-Full}.

\noindent\textbf{Evaluation on BP4D and EmotioNet.} We compared our method with lower-bounds and upper-bounds of the basic model on the test sets of both source domain BP4D and target domain EmotioNet. The F1-frame results of these methods are listed in Table~\ref{tab:comp_baseline}. It can be seen that our method ADLD significantly outperformed the lower-bounds on EmotioNet, in which the margin of average F1-frame is $11.3$ over LI$^{t(l)}_{s(a,l)}$. Without using the pseudo AU labels of the target domain, ADLD still performed better than the upper-bounds on EmotioNet. If the pseudo AU labels are available, our ADLD-Full can achieve the average F1-frame of $42.4$ with a large gap over the upper-bounds. These demonstrate that our method is superior to both lower-bounds and upper-bounds of the basic model for target-domain AU detection.

\noindent\textbf{Evaluation on GFT and EmotioNet.} Table~\ref{tab:comp_baseline_gft} shows the results on the test sets of source domain GFT and target domain EmotioNet. We can observe that our ADLD performed better than both the lower-bounds and the upper-bounds on EmotioNet. Given the target-domain pseudo AU labels, our ADLD-Full further improved the average F1-frame from $41.6$ to $46.0$. Despite being devised for AU detection of the target domain, our ADLD and ADLD-Full also achieved competitive performance on the source domain GFT.

Moreover, there are several interesting observations from Tables~\ref{tab:comp_baseline} and~\ref{tab:comp_baseline_gft}. (i) By using target-domain landmark labels, LI$^{t(l)}_{s(a,l)}$ achieved higher average F1-frame results than LI$_{s(a,l)}$ on both the source domain and the target domain, which indicates that facial landmarks can capture AU-related information to facilitate AU detection. (ii) LI$_{s(a,l)}$ and UI$^{t(a,l)}$ showed bad performance on the target domain and the source domain, respectively. This is because there is a large gap between the constrained source domain and the unconstrained target domain. (iii) UI$^{t(a,l)}_{s(a,l)}$ performed worse than LI$_{s(a,l)}$ on the source domain, and UI$^{t(a,l)}_{s(a,l)}$ also had no apparent advantage over UI$^{t(a,l)}$ on the target domain. This is due to the training of source-domain AU detection and target-domain AU detection would compete against each other without the use of domain transfer.

\begin{table*}
\centering\caption{F1-frame results of our approach and state-of-the-art fully-supervised AU detection methods on the target domain EmotioNet~\cite{benitez2016emotionet} when the source domain datasets are BP4D~\cite{zhang2014bp4d} and GFT~\cite{girard2017sayette}, respectively.}
\label{tab:comp_pseudo}
\begin{tabular}{|*{13}{c|}}
\hline
\multirow{2}*{AU} &\multicolumn{7}{c|}{BP4D+EmotioNet} &\multicolumn{5}{c|}{GFT+EmotioNet}\\
\cline{2-13}&1 &2 &4 &6 &12 &17 &\textbf{Avg} &2 &6 &12 &17 &\textbf{Avg}\\\hline
JAA-Net-I$^{t(a,l)}$ &19.5 &17.5 &46.2 &60.5 &84.9 &14.6 &40.5 &9.0 &64.6 &85.9 &9.0 &42.1\\
JAA-Net-I$^{t(a,l)}_{s(a,l)}$ &21.3 &19.2 &43.6 &59.3 &85.2 &9.3 &39.7 &\textbf{25.5} &62.9 &85.5 &6.8 &45.2\\
ARL-I$^{t(a,l)}$ &23.5 &18.0 &\textbf{49.1} &57.8 &85.2 &\textbf{14.8} &41.4 &18.5 &58.7 &\textbf{86.0} &12.3 &43.9\\
ARL-I$^{t(a,l)}_{s(a,l)}$ &22.3 &17.4 &48.1 &57.5 &84.6 &11.9 &40.3 &15.5 &58.6 &84.3 &\textbf{12.4} &42.7\\
WSC-I$^{t(a,l)}$ &19.8 &14.3 &41.9 &56.6 &\textbf{85.8} &10.5 &38.1 &15.4 &54.5 &85.7 &9.9 &41.4\\
WSC-I$^{t(a,l)}_{s(a,l)}$ &22.7 &16.5 &39.3 &55.2 &83.4 &10.8 &38.0 &16.9 &59.9 &84.7 &9.9 &42.8\\
\textbf{UI}$^{t(a,l)}$ &14.7 &11.4 &41.5 &49.4 &75.8 &11.4 &34.0 &14.8 &53.6 &80.4 &11.0 &39.9\\
\textbf{UI}$^{t(a,l)}_{s(a,l)}$ &15.3 &9.1 &38.5 &48.9 &74.9 &4.4 &31.9 &18.7 &51.7 &80.5 &10.7 &40.4\\
\textbf{ADLD-Full} &\textbf{30.7} &\textbf{26.1} &48.1 &\textbf{60.7} &77.6 &11.5 &\textbf{42.4} &21.9 &\textbf{64.9} &85.4 &11.6 &\textbf{46.0}\\
\hline
\end{tabular}
\end{table*}

\begin{table*}
\centering\caption{F1-frame results of our approach and state-of-the-art fully-supervised AU detection methods on the target domain EmotioNet~\cite{benitez2016emotionet} when the source domain datasets are DISFA~\cite{mavadati2013disfa} and Pain~\cite{lucey2011painful}, respectively.}
\label{tab:comp_pseudo_disfa_pain}
\begin{tabular}{|*{10}{c|}}
\hline
\multirow{2}*{AU} &\multicolumn{6}{c|}{DISFA+EmotioNet} &\multicolumn{3}{c|}{Pain+EmotioNet}\\
\cline{2-10}&4 &6 &12 &25 &26 &\textbf{Avg} &6 &12 &\textbf{Avg}\\\hline
JAA-Net-I$^{t(a,l)}$ &38.7 &60.4 &84.6 &78.8 &\textbf{39.7} &60.4 &58.3 &82.9 &70.6\\
JAA-Net-I$^{t(a,l)}_{s(a,l)}$ &47.0 &59.7 &85.6 &88.1 &34.3 &62.9 &58.4 &84.9 &71.6\\
ARL-I$^{t(a,l)}$ &50.6 &58.6 &85.8 &\textbf{89.8} &34.5 &63.9 &61.2 &85.5 &73.4 \\
ARL-I$^{t(a,l)}_{s(a,l)}$ &\textbf{52.5} &59.6 &86.1 &89.0 &35.5 &64.5 &61.9 &\textbf{85.7} &73.8 \\
\textbf{ADLD-Full} &49.5 &\textbf{63.6} &\textbf{86.8} &88.6 &38.3 &\textbf{65.3} &\textbf{63.6} &\textbf{85.7} &\textbf{74.7} \\
\hline
\end{tabular}
\end{table*}

\begin{table*}
\centering\caption{F1-frame results of our approach and state-of-the-art adversarial domain adaptation methods on the target domain EmotioNet~\cite{benitez2016emotionet} when the source domain datasets are BP4D~\cite{zhang2014bp4d} and GFT~\cite{girard2017sayette}, respectively.}
\label{tab:comp_domain}
\begin{tabular}{|*{13}{c|}}
\hline
\multirow{2}*{AU} &\multicolumn{7}{c|}{BP4D+EmotioNet} &\multicolumn{5}{c|}{GFT+EmotioNet}\\
\cline{2-13}&1 &2 &4 &6 &12 &17 &\textbf{Avg} &2 &6 &12 &17 &\textbf{Avg}\\\hline
DANN-I$^t_{s(a,l)}$ &12.8 &6.9 &18.9 &30.7 &53.1 &6.3 &21.5 &8.9 &34.1 &55.7 &5.1 &25.9\\
DANN-I$^{t(l)}_{s(a,l)}$ &16.8 &12.7 &25.8 &28.9 &62.5 &\textbf{9.5} &26.0 &8.7 &40.3 &63.9 &6.4 &29.8\\
ADDA-I$^t_{s(a,l)}$ &13.8 &6.1 &21.4 &28.5 &57.4 &5.1 &22.1 &11.0 &37.6 &61.0 &5.5 &28.8\\
ADDA-I$^{t(l)}_{s(a,l)}$ &17.7 &5.2 &15.3 &38.2 &58.7 &6.2 &23.6 &9.8 &35.4 &56.7 &6.7 &27.2\\
DRIT-I$^t_{s(a,l)}$ &18.8 &9.0 &27.8 &40.6 &67.9 &5.0 &28.2 &18.1 &48.1 &67.6 &5.6 &34.9\\
DRIT-I$^{t(l)}_{s(a,l)}$ &\textbf{20.4} &7.7 &30.9 &44.2 &67.5 &8.3 &29.8 &\textbf{18.3} &52.3 &73.5 &5.6 &37.4\\
T$^2$Net-I$^t_{s(a,l)}$ &10.1 &5.6 &21.4 &31.3 &57.1 &5.3 &21.8 &6.1 &35.6 &51.7 &6.1 &24.9\\
T$^2$Net-I$^{t(l)}_{s(a,l)}$ &9.4 &9.6 &24.4 &45.1 &69.5 &4.7 &27.1 &15.0 &39.6 &64.5 &6.4 &31.4\\
\textbf{LI}$_{s(a,l)}$ &12.0 &6.9 &11.4 &27.9 &53.5 &3.5 &19.2 &6.0 &32.7 &58.0 &5.2 &25.5\\
\textbf{LI}$^{t(l)}_{s(a,l)}$ &19.0 &8.7 &21.5 &38.1 &58.4 &7.3 &25.5 &8.6 &44.2 &69.2 &6.4 &32.1\\
\textbf{ADLD} &19.8 &\textbf{25.2} &\textbf{31.0} &\textbf{58.2} &\textbf{78.3} &8.6 &\textbf{36.8} &17.4 &\textbf{59.3} &\textbf{80.2} &\textbf{9.5} &\textbf{41.6}\\
\hline
\end{tabular}
\end{table*}

\subsection{Comparison with State-of-the-Art Methods}

We compared our approach against state-of-the-art methods, including fully-supervised AU detection methods using pseudo AU labels and adversarial domain adaptation methods. All methods compared were implemented with their released code.

\subsubsection{Fully-Supervised AU Detection}

To enable the comparison with fully-supervised AU detection methods, we considered the scenario where target-domain pseudo AU labels are available. For a reliable comparison, we only compared state-of-the-art AU detection methods with code released. Recently, there are two fully-supervised AU detection methods JAA-Net~\cite{shao2018deep} and ARL~\cite{shao2019facial}, as well as an AU detection technique WSC~\cite{zhao2018learning} specialized for inaccurate pseudo AU labels.

Specifically, we trained JAA-Net and ARL by using landmark labels and pseudo AU labels of the target domain, and obtained \textbf{JAA-Net-I}$^{t(a,l)}$ and \textbf{ARL-I}$^{t(a,l)}$, respectively. Note that ARL does not require landmarks, so landmark labels were actually not used. By further using source-domain landmark and AU labels, we can obtain \textbf{JAA-Net-I}$^{t(a,l)}_{s(a,l)}$ and \textbf{ARL-I}$^{t(a,l)}_{s(a,l)}$. WSC exploits AU-related features to refine the pseudo AU labels, and then uses the re-annotated AU labels to retrain AU detection. We employed UI$^{t(a,l)}_{s(a,l)}$ and UI$^{t(a,l)}$ to extract AU-related features respectively, in which the output of the global average pooling layer of each branch in $F_a$ is treated as a related feature for the corresponding AU. This follows the setting of WSC that each AU is processed independently. With the target-domain landmark labels and re-annotated AU labels, we can further retrain UI$^{t(a,l)}_{s(a,l)}$ and UI$^{t(a,l)}$ by adopting and not adopting source-domain images, which are denoted as \textbf{WSC-I}$^{t(a,l)}_{s(a,l)}$ and \textbf{WSC-I}$^{t(a,l)}$, respectively.

Table~\ref{tab:comp_pseudo} shows the F1-frame results of our ADLD-Full and state-of-the-art fully-supervised AU detection methods in the scenario with target-domain pseudo AU labels. It can be seen that our method outperformed previous fully-supervised AU detection methods on EmotioNet for any one source domain dataset. Note that JAA-Net-I$^{t(a,l)}$, JAA-Net-I$^{t(a,l)}_{s(a,l)}$, ARL-I$^{t(a,l)}$ and ARL-I$^{t(a,l)}_{s(a,l)}$ performed significantly better than the upper-bounds UI$^{t(a,l)}$ and UI$^{t(a,l)}_{s(a,l)}$. This is because our AU detector $F_a$ has a less complex structure than the state-of-the-art JAA-Net and ARL. Our main goal is to propose an effective AU label transfer method rather than a complex fully-supervised AU detector. With a less complex $F_a$, our ADLD-Full still achieved better performance than JAA-Net and ARL. Besides, although WSC can refine the inaccurate pseudo AU labels, its results are worse than our ADLD-Full which transfers accurate AU labels from the source domain.

We also show the comparison results in Table~\ref{tab:comp_pseudo_disfa_pain} when the source domain datasets are DISFA and Pain. It can be observed that our ADLD-Full still achieved the highest average F1-frame of $65.3$ and $74.7$ for DISFA and Pain, respectively. This demonstrates that our framework works well in the scenario of using target-domain pseudo AU labels.


\begin{table*}
\centering\caption{F1-frame results of our approach and state-of-the-art adversarial domain adaptation methods on the target domain EmotioNet~\cite{benitez2016emotionet} when the source domain datasets are DISFA~\cite{mavadati2013disfa} and Pain~\cite{lucey2011painful}, respectively.}
\label{tab:comp_domain_disfa_pain}
\begin{tabular}{|*{10}{c|}}
\hline
\multirow{2}*{AU} &\multicolumn{6}{c|}{DISFA+EmotioNet} &\multicolumn{3}{c|}{Pain+EmotioNet}\\
\cline{2-10}&4 &6 &12 &25 &26 &\textbf{Avg} &6 &12 &\textbf{Avg}\\\hline
DRIT-I$^t_{s(a,l)}$ &\textbf{39.7} &40.7 &69.2 &74.9 &17.8 &48.5 &36.2 &73.8 &55.0\\
DRIT-I$^{t(l)}_{s(a,l)}$ &38.0 &49.0 &64.4 &76.4 &19.2 &49.4 &37.0 &\textbf{74.8} &55.9 \\
T$^2$Net-I$^t_{s(a,l)}$ &19.6 &31.2 &52.1 &68.6 &22.5 &38.8 &38.0 &46.1 &42.0 \\
T$^2$Net-I$^{t(l)}_{s(a,l)}$ &23.0 &37.9 &61.6 &80.9 &28.0 &46.3 &45.2 &51.3 &48.3 \\
\textbf{ADLD} &27.0 &\textbf{51.8} &\textbf{73.8} &\textbf{88.4} &\textbf{34.2} &\textbf{55.0} &\textbf{52.8} &69.1 &\textbf{60.9}\\
\hline
\end{tabular}
\end{table*}

\begin{table*}
\centering\caption{F1-frame results of our approach and state-of-the-art adversarial domain adaptation methods on the target domain VGGFace2~\cite{cao2018vggface2} when the source domain datasets are BP4D~\cite{zhang2014bp4d} and GFT~\cite{girard2017sayette}, respectively.}
\label{tab:comp_domain_vggface2}
\begin{tabular}{|*{13}{c|}}
\hline
\multirow{2}*{AU} &\multicolumn{7}{c|}{BP4D+VGGFace2} &\multicolumn{5}{c|}{GFT+VGGFace2}\\
\cline{2-13}&1 &2 &4 &6 &12 &17 &\textbf{Avg} &2 &6 &12 &17 &\textbf{Avg}\\\hline
DRIT-I$^t_{s(a,l)}$ &21.2 &16.0 &30.8 &43.0 &69.7 &5.7 &31.1 &17.1 &48.4 &66.9 &6.2 &34.6\\
DRIT-I$^{t(l)}_{s(a,l)}$ &\textbf{21.8} &22.3 &\textbf{36.6} &44.1 &63.9 &8.2 &32.8 &18.6 &52.9 &68.1 &7.3 &36.7\\
T$^2$Net-I$^t_{s(a,l)}$ &8.8 &7.5 &17.4 &31.4 &57.7 &4.6 &21.2 &6.1 &35.6 &51.7 &6.1 &24.9\\
T$^2$Net-I$^{t(l)}_{s(a,l)}$ &20.8 &14.4 &18.3 &43.3 &69.1 &4.4 &28.4 &17.0 &34.6 &65.6 &\textbf{14.6} &32.9\\
\textbf{ADLD} &19.7 &\textbf{25.0} &27.4 &\textbf{54.8} &\textbf{75.9} &\textbf{10.0} &\textbf{35.5} &\textbf{19.7} &\textbf{60.4} &\textbf{76.3} &9.0 &\textbf{41.4}\\
\hline
\end{tabular}
\end{table*}

\begin{table*}
\centering\caption{F1-frame results for different variants of our ADLD on the target domain EmotioNet~\cite{benitez2016emotionet} when the source domain dataset is BP4D~\cite{zhang2014bp4d}. Except for ADLD (input $x^t$), other methods input $G(z^t_l,z^t_t)$ to $F_a$ to predict the AU occurrence probabilities at test time, as illustrated in Fig.~\ref{fig:ADLD_framework_full}(c).}
\label{tab:variant_f1}
\begin{tabular}{|c|*{5}{c}*{8}{c|}}
\hline
\multirow{2}*{Method} &\multirow{2}*{$\mathcal{L}_{a}$} &\multirow{2}*{$\mathcal{L}_{l}$}
&\multirow{2}*{$\mathcal{L}_{r}$} &\multirow{2}*{$\mathcal{L}_{cc}$} &\multirow{2}*{$\mathcal{L}_{{ad}_l}$}
&\multirow{2}*{$\mathcal{L}_{{ad}_f}$} &\multicolumn{7}{c|}{AU}\\\cline{8-14}
& & & & & & &1 &2 &4 &6 &12 &17 &\textbf{Avg}\\
\hline
B-Net &$\surd$ &$\surd$ & & & & &12.1 &14.3 &\textbf{37.6} &38.4 &53.1 &4.6 &26.7\\
B-Net+$\mathcal{L}_{r}$ &$\surd$ &$\surd$ &$\surd$ & & & &14.8 &16.6 &29.5 &47.2 &64.6 & 3.1 &29.3\\
B-Net+$\mathcal{L}_{r}$+$\mathcal{L}_{cc}$ &$\surd$ &$\surd$ &$\surd$ &$\surd$ & & &19.6 &21.8 &24.2 &40.1 &75.3 &5.9 &31.2\\
B-Net+$\mathcal{L}_{r}$+$\mathcal{L}_{cc}$+$\mathcal{L}_{{ad}_l}$ &$\surd$ &$\surd$ &$\surd$ &$\surd$ &$\surd$ & &17.6 &20.9 &36.4 &47.4 &77.4 &6.2 &34.3\\
\textbf{ADLD} &$\surd$ &$\surd$ &$\surd$ &$\surd$ &$\surd$ &$\surd$ &19.8 &\textbf{25.2} &31.0 &\textbf{58.2} &\textbf{78.3} &\textbf{8.6} &\textbf{36.8}\\
\textbf{ADLD} (input $x^t$) &$\surd$ &$\surd$ &$\surd$ &$\surd$ &$\surd$ &$\surd$ &\textbf{20.2} &10.3 &25.2 &30.0 &62.4 &7.1 &25.8\\
\hline
\end{tabular}
\end{table*}

\subsubsection{Adversarial Domain Adaptation}

To evaluate the effectiveness of AU label transfer, we compared our ADLD with typical adversarial domain adaptation methods. These methods include DANN~\cite{ganin2016domain} and ADDA~\cite{tzeng2017adversarial} which learn domain-invariant features, and DRIT~\cite{lee2018diverse} and T$^2$Net~\cite{zheng2018t2net} which translate the source images into target-style images. For a fair comparison, an AU detection network with the same structure as the basic model was applied to these methods.

Particularly, for DANN and ADDA, $E_f$ is encouraged to learn a domain-invariant rich feature by a domain discriminator with the same structure as $D^s_f$. We implemented DANN and ADDA by employing and not employing target-domain landmark labels, respectively. Taking DANN as an example, we denote it as \textbf{DANN-I}$^{t(l)}_{s(a,l)}$ and \textbf{DANN-I}$^t_{s(a,l)}$, respectively. For DRIT, we used its original framework architecture to generate target-style images by transferring the style from the target images to the source images. Then we used the generated target-style images with inherited AU and landmark labels to train AU detection, in which we similarly obtained two variants \textbf{DRIT-I}$^{t(l)}_{s(a,l)}$ and \textbf{DRIT-I}$^t_{s(a,l)}$. We applied the same setting of DRIT to T$^2$Net, except we simultaneously trained the image translation network and the AU detection network, following the original setting of T$^2$Net.

\noindent\textbf{Evaluation on Target Domain EmotioNet.} Tables~\ref{tab:comp_domain} and~\ref{tab:comp_domain_disfa_pain} summarize the F1-frame results of these methods on EmotioNet. We can see that our method ADLD remarkably outperformed the state-of-the-art adversarial domain adaptation methods, including both the domain-invariant feature based and image translation based methods. Compared to the average F1-frame $(19.2,25.5)$ of the lower-bound LI$_{s(a,l)}$ for the source domain BP4D and GFT, DANN-I$^t_{s(a,l)}$ and ADDA-I$^t_{s(a,l)}$ only improved with small margins of $(2.3,0.4)$ and $(2.9,3.3)$ respectively by using target-domain training images. Besides, DANN-I$^{t(l)}_{s(a,l)}$ and ADDA-I$^{t(l)}_{s(a,l)}$ overall performed worse than the lower-bound LI$^{t(l)}_{s(a,l)}$. This is because enforcing domain invariance of features for inputting to the AU detector $F_a$ may neglect AU-related information.

Moreover, DRIT-I$^t_{s(a,l)}$, DRIT-I$^{t(l)}_{s(a,l)}$, T$^2$Net-I$^t_{s(a,l)}$ and T$^2$Net-I$^{t(l)}_{s(a,l)}$ all performed much worse than our ADLD in both Tables~\ref{tab:comp_domain} and~\ref{tab:comp_domain_disfa_pain}. This demonstrates that only translating the image style has a limited contribution to the target-domain AU detection, since major domain shifts including distribution variations of pose and occlusion are not reduced. In contrast, our method alleviates such problems by mapping images to a latent feature domain specialized for the target-domain AU detection.

\noindent\textbf{Evaluation on Target Domain VGGFace2.} Table~\ref{tab:comp_domain_vggface2} shows the comparison results on the target domain VGGFace2. It can be seen that our ADLD significantly outperformed other adversarial domain adaptation methods for both source domain datasets BP4D and GFT. Note that VGGFace2 shares the same testing set with EmotioNet, so there is a domain gap between the training and testing sets of VGGFace2. In this challenging case, our ADLD still achieved similar performance to the evaulation on EmotioNet in Table~\ref{tab:comp_domain}.

\begin{figure*}
\centering\includegraphics[width=\linewidth]{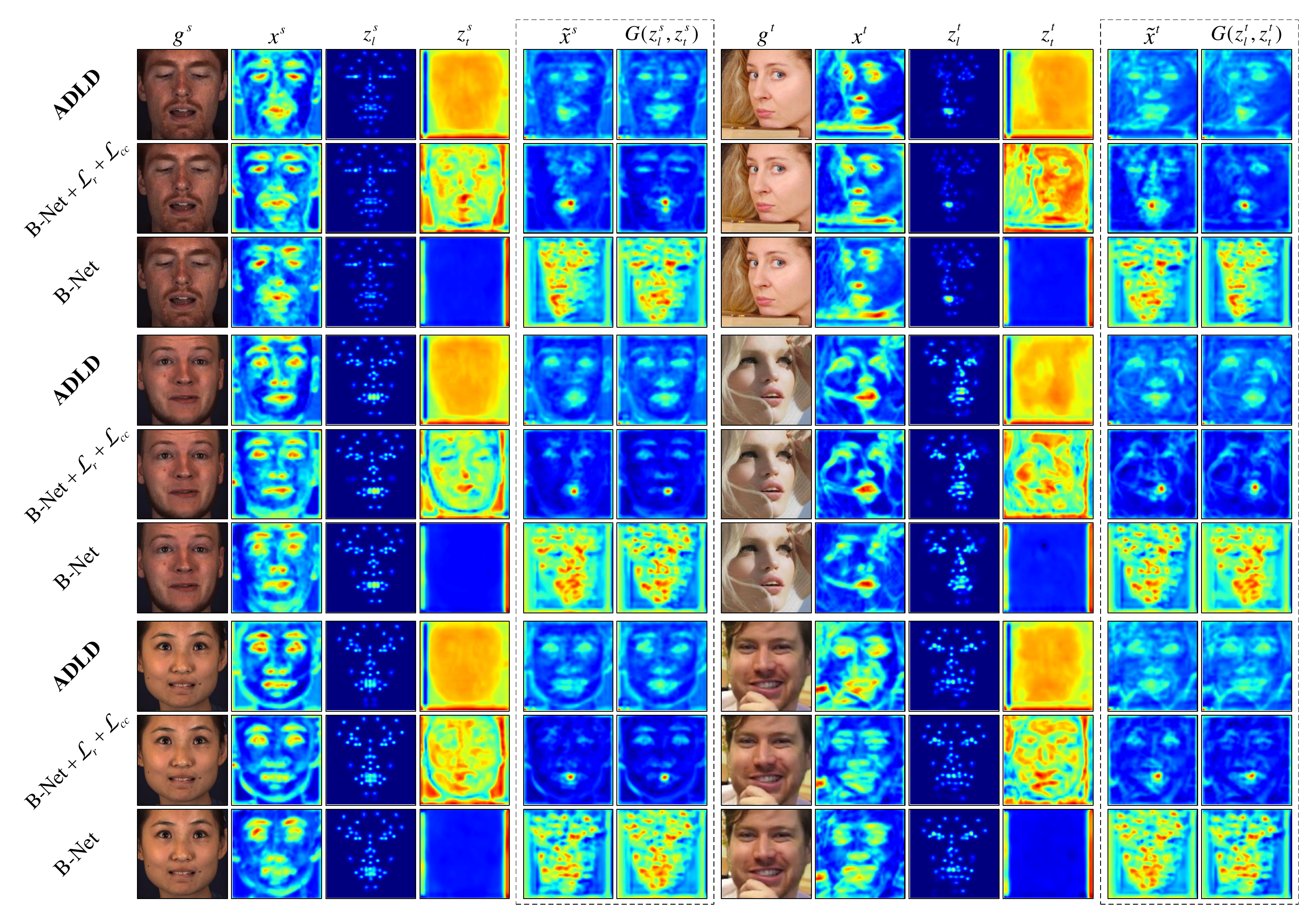}
\caption{Visualization of features for B-Net, B-Net+$\mathcal{L}_{r}$+$\mathcal{L}_{cc}$ and our ADLD with three input pairs of source BP4D~\cite{zhang2014bp4d} and target EmotioNet~\cite{benitez2016emotionet} images. Compared to the source images $g^s$, the target images $g^t$ have different expressions and poses, and may be partially occluded. \textit{$x^s$ and $x^t$ are rich features, $z^s_l$ and $z^t_l$ are landmark-related features, $z^s_t$ and $z^t_t$ are landmark-free features, $\tilde{x}^s$ and $\tilde{x}^t$ are latent features, and $G(z^s_l,z^s_t)$ and $G(z^t_l,z^t_t)$ are self-reconstructed latent
features.} $\tilde{x}^s$, $\tilde{x}^t$, $G(z^s_l,z^s_t)$ and $G(z^t_l,z^t_t)$ from the latent feature domain are shown in the dotted boxes. We expect $\tilde{x}^t$ to contain preserved global pose and texture from $x^t$, and transferred AU-related inner-landmark information from $x^s$.}
\label{fig:visualize_feature}
\end{figure*}

\subsection{Ablation Study}
In this section, we study the effectiveness of main loss terms in Eq.~\eqref{eq:full} for our framework. Table~\ref{tab:variant_f1} summarizes the structures and F1-frame results of different variants of our ADLD on EmotioNet. Fig.~\ref{fig:visualize_feature} visualizes the features of ADLD and its variants for three example pairs of input images, in which the unpaired source and target images exhibit different expressions, poses, illuminations, occlusions and backgrounds.

\subsubsection{B-Net with $\mathcal{L}_a$ and $\mathcal{L}_l$}

The baseline network B-Net uses the same architecture as ADLD with only the losses $\mathcal{L}_a$ and $\mathcal{L}_l$. It can be observed that B-Net failed to achieve good performance, in which its average F1-frame is just slightly higher than $25.5$ of the lower-bound LI$^{t(l)}_{s(a,l)}$. compared to the self-reconstructed latent feature $G(z^t_l,z^t_t)$, the latent feature $\tilde{x}^t$ of B-Net cannot preserve target-domain AU-free information like facial global pose. This is because the landmark-free feature $z^t_t$ just simply removes all facial shape information including both inner-landmarks and global pose, without constraints from other losses. In this case, $\tilde{x}^t$ is similar to $G(z^s_l,z^s_t)$, and is effective for AU detection of the source domain rather than the target domain, which results in the low performance on EmotioNet.

Note that the landmark-related feature $z^t_l$ which highlights the locations of landmarks is adaptively obtained. If the $i$-th landmark is difficult to localize, the response $\sigma(F_{l}^{(i,y^t_i)}(x^t))$ at its location $(\lfloor q^t_{2i-1}d/l\rceil, \lfloor q^t_{2i}d/l\rceil)$ will not be significantly higher than other locations on its response map. By element-wise summing the response maps of all landmarks in Eq.~\eqref{eq:shape},
our learned $z^t_l$ can obtain wider responses around a landmark that is more difficult to localize, so as to capture more information to alleviate the influence of inaccurate localization. Another possible way is to manually generate a response map as the landmark-related feature, in which a predefined Gaussian distribution is used to generate responses around the predicted location of each landmark. The landmarks with different localization difficulties are treated equally, which may cause the loss of useful information for challenging landmarks.

\subsubsection{$\mathcal{L}_{r}$ and $\mathcal{L}_{cc}$}

In our framework, the self-reconstruction loss $\mathcal{L}_{r}$ and the cross-cycle consistency loss $\mathcal{L}_{cc}$ are introduced for constraining the mapping from source and target domains to the latent feature domain. It can be observed from Table \ref{tab:variant_f1} that B-Net+$\mathcal{L}_{r}$ increased the average F1-frame to $29.3$ over B-Net. After applying $\mathcal{L}_{cc}$, the result was further improved to $31.2$.

Due to the supervisions of $\mathcal{L}_{r}$ and $\mathcal{L}_{cc}$, in Fig.~\ref{fig:visualize_feature} we can see that the learned $\tilde{x}^t$ of B-Net+$\mathcal{L}_{r}$+$\mathcal{L}_{cc}$ can coarsely inherit source-domain AU-related inner-landmark information and preserve target-domain AU-free global pose. However, facial global contour and inner shape of $\tilde{x}^t$ are not very clear. This is because the use of only $\mathcal{L}_{a}$, $\mathcal{L}_{l}$, $\mathcal{L}_{r}$ and $\mathcal{L}_{cc}$ cannot effectively enforce $z^t_t$ to discard inner-landmark information and keep global pose. In this case, the learned latent feature domain has limited effectiveness for target-domain AU detection.

\subsubsection{$\mathcal{L}_{{ad}_l}$ and $\mathcal{L}_{{ad}_f}$}

After adding the landmark adversarial loss $\mathcal{L}_{{ad}_l}$ for the landmark-free features, the average F1-frame was improved from $31.2$ to $34.3$. This profits from $\mathcal{L}_{{ad}_l}$ which adversarially disentangles facial inner-landmark information for $z^t_t$. When further using the feature adversarial loss $\mathcal{L}_{{ad}_f}$, our ADLD achieved the best performance. $\mathcal{L}_{{ad}_f}$ is beneficial for the latent feature $\tilde{x}^t$ to preserve target-domain global pose and texture information from $x^t$.

In Fig.~\ref{fig:visualize_feature}, we can observe that the facial inner shape of $\tilde{x}^t$ is similar to those of $G(z^s_l, z^s_t)$, and the facial global contour of $\tilde{x}^t$ is similar to that of $G(z^t_l, z^t_t)$. This demonstrates that the learned latent feature domain can preserve source-domain AU-related information and target-domain AU-free information, which is specialized for target-domain AU detection. With the latent feature domain, our method can exploit available and accurate source-domain AU labels and adapt to unconstrained conditions of the target domain such as large poses, partial occlusions and arbitrary backgrounds.

\subsubsection{Latent Feature Domain for AU Detection}

It can be seen from Fig.~\ref{fig:visualize_feature} that the latent feature domain has a similar structure to the domain of rich features, but with different details. If we directly input $x^t$ to the AU detector $F_a$ in Fig.~\ref{fig:ADLD_framework_full}(c), our ADLD only achieved the average F1-frame of $25.8$, much worse than $36.8$ of using $G(z^t_l, z^t_t)$. This demonstrates that the latent feature domain is not just obtained by a simple domain mapping, but instead is learned by disentangling landmark-free and landmark-related features and maximizing the performance of target-domain AU detection.

Moreover, since there are large gaps like pose differences between the constrained source domain and the unconstrained target domain, it is difficult to integrate the information from different domains into a realistic image. The latent feature domain has a larger capacity to combine landmark-related information with target-domain landmark-free information than the image domain. Besides, our goal is to achieve target-domain AU detection instead of synthesizing images. Image generation requires more complex network structures than feature translation, as each image pixel needs numerous computations.

\subsection{Validation of Landmark Definition}

\begin{figure}
\centering\includegraphics[width=\linewidth]{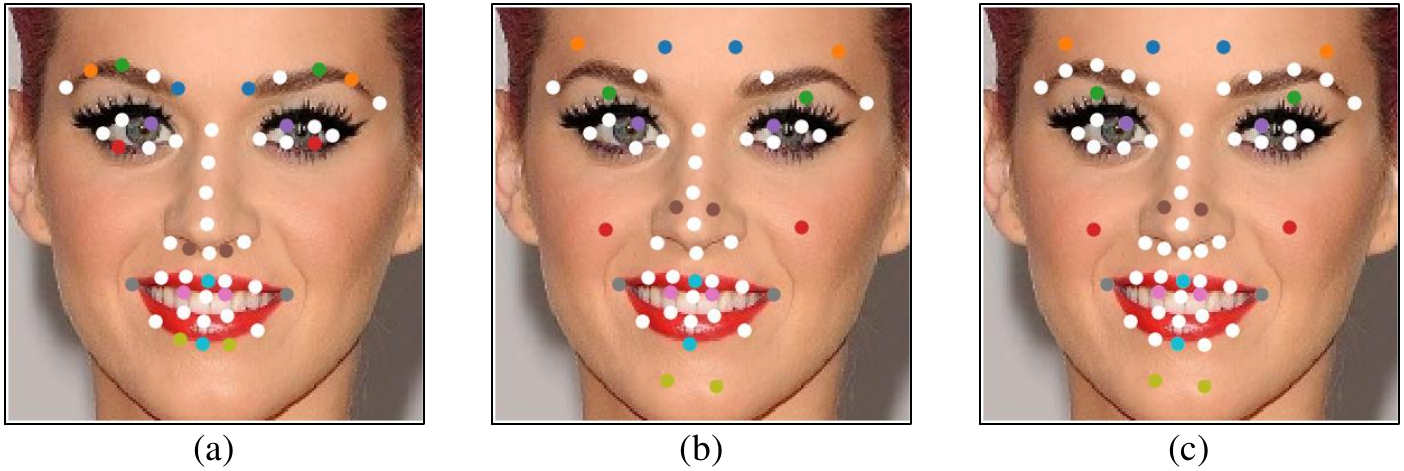}
\caption{Different types of landmark definitions. \textbf{(a)} Original definition. \textbf{(b)} New definition by replacing the corresponding landmarks using the predefined AU centers. \textbf{(c)} New definition by adding the predefined AU centers.}
\label{fig:landmark_all_definition}
\end{figure}

To evaluate the effectiveness of our landmark definition in Fig.~\ref{fig:landmark_definition}(b), we implemented a variant of our approach using the original landmark definition in Fig.~\ref{fig:landmark_definition}(a). Since an alternative solution of defining AU-related landmarks is to add the predefined AU centers into the original landmark definition, we implemented another variant of our approach using this new definition. These three types of landmark definitions are illustrated in Fig.~\ref{fig:landmark_all_definition}. We show the F1-frame results of our approach using different landmark definitions in Table~\ref{tab:res_different_lands}.

It can be observed that our ADLD and ADLD-Full using the landmark definition in Fig.~\ref{fig:landmark_all_definition}(b) both outperformed the variants using other two landmark definitions. This is because the original landmark definition in Fig.~\ref{fig:landmark_all_definition}(a) fails to capture accurate AU-related information. On the other hand, the landmark definition in Fig.~\ref{fig:landmark_all_definition}(c) has redundant landmark information and thus limits the capability of focusing on AU-related information. Our proposed landmark definition in Fig.~\ref{fig:landmark_all_definition}(b) is beneficial for capturing the most related AU information so as to facilitate the target-domain AU detection.

We also notice that the improvements of ADLD-Full over ADLD-Full$_{ori}$ and ADLD-Full$_{add}$ are smaller than the improvements of ADLD over ADLD$_{ori}$ and ADLD$_{add}$. The main difference between ADLD-Full and ADLD is the former has target-domain pseudo AU labels. Additionally, landmark definition is related to the effectiveness of AU label transfer. Since the effect from pseudo AU labels and the effect from AU label transfer are integrated in ADLD-Full, the improvement from AU label transfer may be degraded. This also causes the performances with the landmark definitions in Fig.~\ref{fig:landmark_all_definition}(a) and Fig.~\ref{fig:landmark_all_definition}(c) fluctuated in ADLD-Full and ADLD. We think the experimental results on ADLD is more convincing, and the landmark definition in Fig.~\ref{fig:landmark_all_definition}(c) is potentially better than the landmark definition in Fig.~5(a) for AU label transfer.

\begin{table}
\centering\caption{F1-frame results of our approach using different landmark definitions on the target domain EmotioNet~\cite{benitez2016emotionet} when the source domain dataset is BP4D~\cite{zhang2014bp4d}. Taking ADLD as an example, we denote its variants using landmark definitions in Fig.~\ref{fig:landmark_all_definition}(a) and (c) as ADLD$_{ori}$ and ADLD$_{add}$, respectively.}
\label{tab:res_different_lands}
\begin{tabular}{|*{8}{c|}}
\hline
AU &1 &2 &4 &6 &12 &17 &Avg\\
\hline
ADLD$_{ori}$ &20.0 &21.2 &\textbf{35.9} &44.5 &66.6 &9.6 &33.0\\
ADLD$_{add}$ &\textbf{20.5} &15.8 &33.6 &56.5 &70.7 &\textbf{10.0} &34.5\\
\textbf{ADLD} &19.8 &\textbf{25.2} &31.0 &\textbf{58.2} &\textbf{78.3} &8.6 &\textbf{36.8}\\
\hline
ADLD-Full$_{ori}$ &27.3 &\textbf{30.5} &36.9 &58.4 &\textbf{83.8} &\textbf{16.3} &42.2\\
ADLD-Full$_{add}$ &26.9 &16.5 &42.2 &59.6 &82.2 &14.4 &40.3\\
\textbf{ADLD-Full} &\textbf{30.7} &26.1 &\textbf{48.1} &\textbf{60.7} &77.6 &11.5 &\textbf{42.4}\\
\hline
\end{tabular}
\end{table}

\begin{table*}
\centering\caption{F1-frame results of different AU detectors when the source domain dataset is BP4D~\cite{zhang2014bp4d} or GFT~\cite{girard2017sayette} and the target domain dataset is EmotioNet~\cite{benitez2016emotionet}. ``Source'' and ``Target'' denote the results on the test sets of source domain and target domain, respectively. Since source-domain landmark-related feature is transferred to the target domain in our ADLD framework, we also show the results of $F_{a_{(l)}}$ on the target domain.}
\label{tab:au_detectors_res}
\begin{tabular}{|*{14}{c|}}
\hline
\multicolumn{2}{|c|}{\multirow{2}{*}{AU}} &\multicolumn{7}{c|}{BP4D+EmotioNet} &\multicolumn{5}{c|}{GFT+EmotioNet}\\
\cline{3-14} \multicolumn{2}{| c|}{}&1 &2 &4 &6 &12 &17 &\textbf{Avg} &2 &6 &12 &17 &\textbf{Avg}\\\hline
\multirow{4}*{Source} &$F_{a_{(none)}}$ &0 &0 &0 &0 &69.2 &0 &11.5 &0 &0 &45.2 &0 &11.3\\
&$F_{a_{(t)}}$ &14.1 &7.8 &17.6 &38.6 &37.2 &28.5 &24.0 &12.8 &42.7 &36.2 &41.8 &33.4\\
&$F_{a_{(l)}}$ &47.0 &5.4 &50.7 &70.0 &\textbf{79.6} &56.4 &51.5 &39.7 &68.0 &68.2 &51.0 &56.7\\
&ADLD &\textbf{50.5} &\textbf{35.7} &\textbf{61.8} &\textbf{74.1} &75.2 &\textbf{69.0} &\textbf{61.0} &\textbf{39.8} &\textbf{79.3} &\textbf{81.4} &\textbf{54.9} &\textbf{63.8}\\\hline
\multirow{2}*{Target} &$F_{a_{(l)}}$ &\textbf{21.2} &16.9 &29.7 &42.9 &55.6 &6.6 &28.8 &9.3 &43.0 &72.0 &9.3 &33.4\\
&ADLD &19.8 &\textbf{25.2} &\textbf{31.0} &\textbf{58.2} &\textbf{78.3} &\textbf{8.6} &\textbf{36.8} &\textbf{17.4} &\textbf{59.3} &\textbf{80.2} &\textbf{9.5} &\textbf{41.6}\\
\hline
\end{tabular}
\end{table*}

\subsection{Limitations}

The disentanglement of AU-related information and AU-free information is a challenging issue in domain adaptation based AU detection. It is hard to do a perfect disentanglement. In this section, we analyzed the amount of AU information in the disentangled landmark-free feature and landmark-related feature. Since only the source domain has accurate AU labels, we used the trained ADLD model to extract landmark-free features and landmark-related features of source-domain training samples, and then retrained the AU detector $F_a$ by inputting the landmark-free feature and the landmark-related feature, denoted as $F_{a_{(t)}}$ and $F_{a_{(l)}}$, respectively. To evaluate whether the two AU detectors work better than a random guess, we also implemented $F_{a_{(none)}}$ as the baseline by inputting a zero-value feature to $F_a$. Their AU detection results are presented in Table~\ref{tab:au_detectors_res}.

We can see that the results of $F_{a_{(t)}}$ were better than those of the random guess of $F_{a_{(none)}}$, and were significantly worse than those of $F_{a_{(l)}}$ on the source domain. This demonstrates that the landmark-free feature does contain a little AU information. 
However, it is better to remove the source-domain landmark-free feature since the performance damage it causes in the target domain, due to the domain gap, overtakes the little AU information it contains. On the contrary, it is better to add the target-domain landmark-free feature since the target-domain texture context it brings in might well surpass the mismatched issue. This can be seen from the margin between the results of $F_{a_{(l)}}$ and ADLD.
In our ADLD framework, we combine source-domain landmark-related feature with target-domain landmark-free feature in the latent feature domain, in which the latter can supplement useful domain-related texture information for target-domain AU detection. 

\section{Conclusion}

In this paper, we proposed an end-to-end unconstrained facial AU detection framework by transferring the available and accurate AU labels from the constrained source domain to the unconstrained target domain. We proposed to map the source and target domains to a latent feature domain which is specialized for the target-domain AU detection. To achieve the domain mapping, we also proposed a novel landmark adversarial loss to disentangle the landmark-free feature and the landmark-related feature. Moreover, our framework can be naturally extended to the scenario with target-domain pseudo AU labels.

We compared our proposed framework with two lower-bounds and two upper-bounds of the basic model on the challenging benchmarks. The experimental results demonstrated that our framework soundly outperforms both the lower-bounds and the upper-bounds. In addition, we compared our method against state-of-the-art approaches involving fully-supervised AU detection methods using target-domain pseudo AU labels and adversarial domain adaptation methods. The results showed that our method performs better than all these previous works. We also conducted an ablation study which indicates that the loss terms in our framework are effective, and the learned latent feature domain combining source-domain AU-related information with target-domain AU-free information is beneficial for the target-domain AU detection.

We further conducted experiments to validate that our proposed landmark definition is beneficial for AU detection. Our method can be generalized as mapping source and target domains to a latent feature domain where source task-related feature and target task-free feature are combined, by maximizing the performance of target-domain task. We believe this idea is also promising for other domain adaption problems.



%

%

\ifCLASSOPTIONcompsoc
  \section*{Acknowledgments}
\else
  \section*{Acknowledgment}
\fi

This work is partially supported by the National Key R\&D Program of China (No. 2019YFC1521104), the National Natural Science Foundation of China (No. 61972157), the Natural Science Foundation of Jiangsu Province (No. BK20201346), the Six Talent Peaks Project in Jiangsu Province (No. 2015-DZXX-010), the Zhejiang Lab (No. 2020NB0AB01), the Data Science \& Artificial Intelligence Research Centre@NTU (DSAIR), the Monash FIT Start-up Grant, and the Fundamental Research Funds for the Central Universities (No. 2021QN1072).

\ifCLASSOPTIONcaptionsoff
  \newpage
\fi



\bibliographystyle{IEEEtran}
\bibliography{references}

%
\begin{IEEEbiography}[{\includegraphics[width=1in,height=1.25in,clip,keepaspectratio]{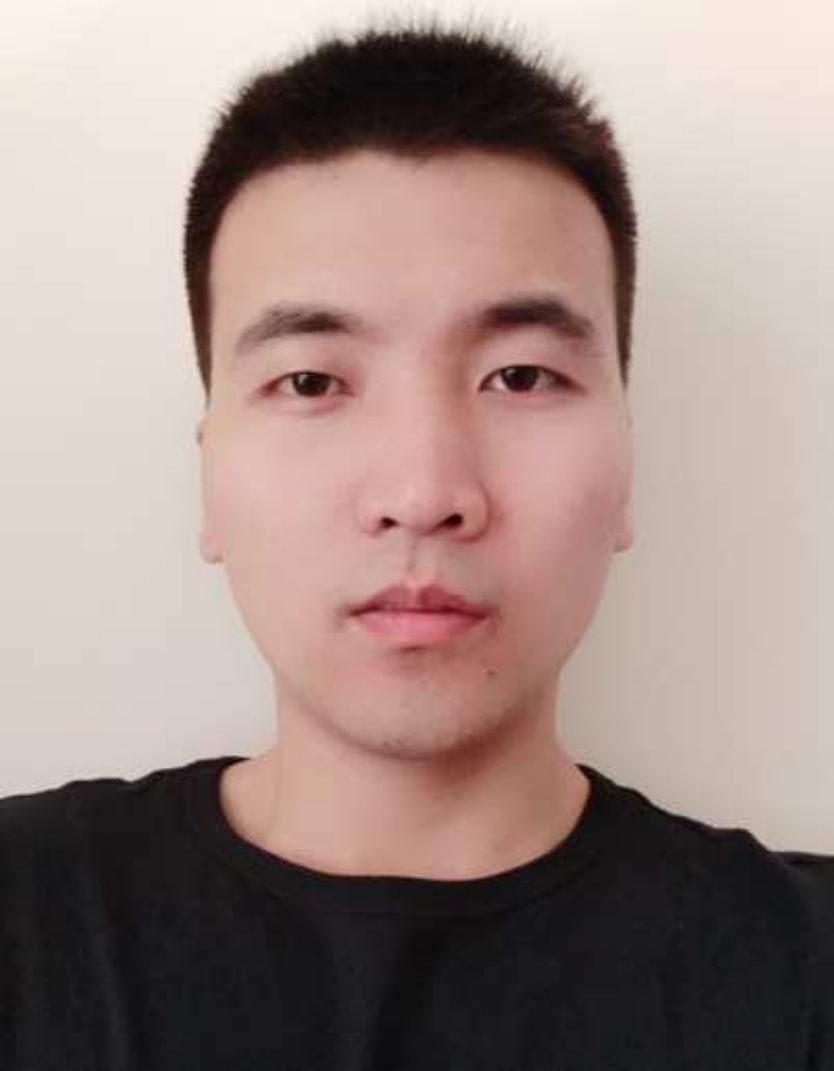}}]{Zhiwen Shao}
received his B.Eng. degree in Computer Science and Technology from the Northwestern Polytechnical University, China in 2015. He received the Ph.D. degree from the Shanghai Jiao Tong University, China in 2020. He is now a Tenure-Track Associate Professor at the School of Computer Science and Technology, China University of Mining and Technology, China. From 2017 to 2018, he was a joint Ph.D. student at the Multimedia and Interactive Computing Lab, Nanyang Technological University, Singapore. His research interests lie in face analysis and deep learning, in particular, facial expression recognition, facial expression manipulation, and face alignment. He has served as a PC member in AAAI 2021 and IJCAI 2021.
\end{IEEEbiography}

\begin{IEEEbiography}[{\includegraphics[width=1in,height=1.25in,clip,keepaspectratio]{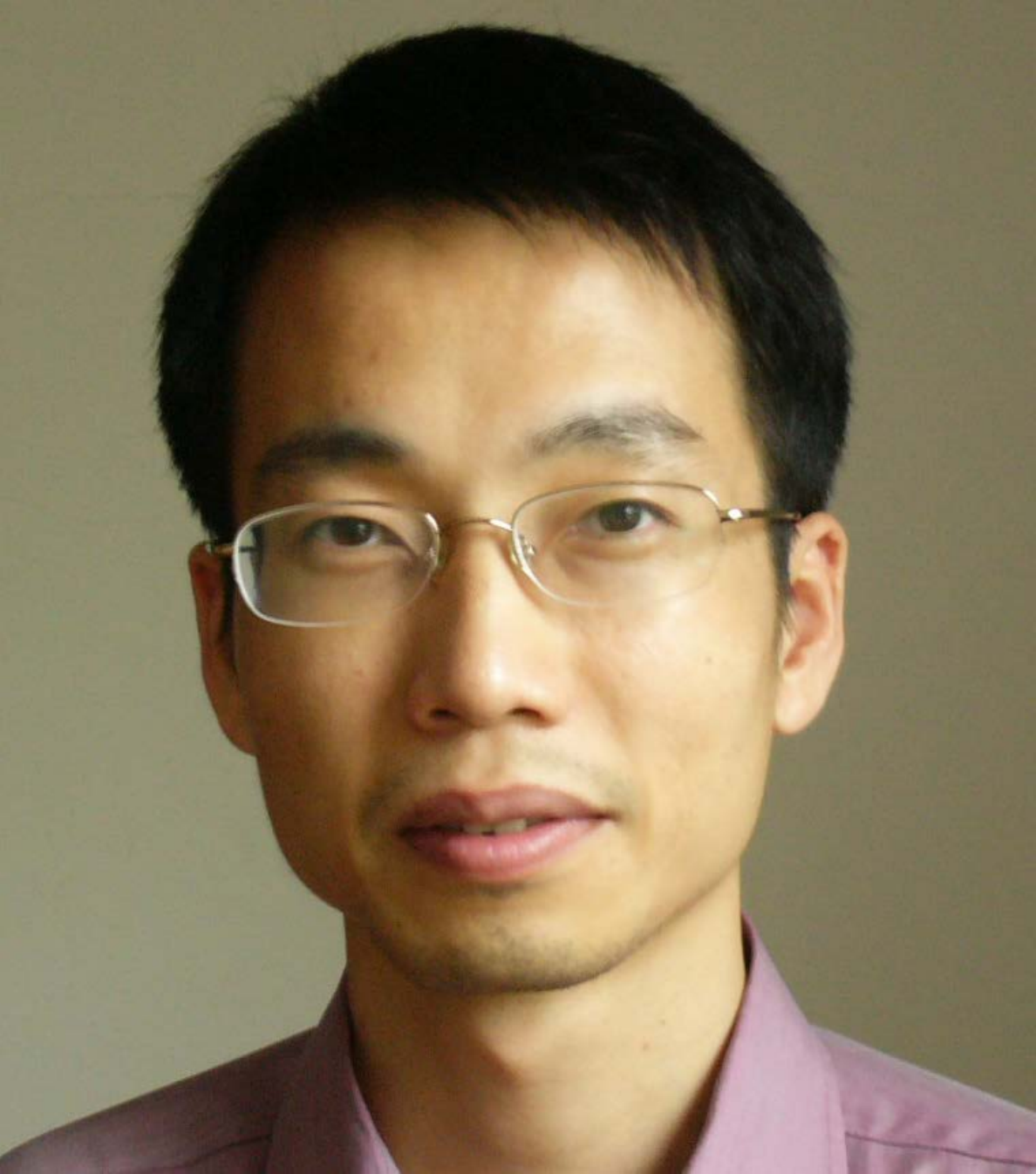}}]{Jianfei Cai}
received his Ph.D. degree from the University of Missouri-Columbia in 2002. He is currently a Full Professor and has served as the Group Lead for the Data Science \& Artificial Intelligence Group at the Faculty of Information Technology, Monash University, Australia. Before that, he was a Full Professor and a Cluster Deputy Director of Data Science \& Artificial Intelligence Research Centre at the Nanyang Technological University, Singapore. He has published over 200 technical papers in international journals and conferences. His major research interests include computer vision, multimedia and deep learning. He is an IEEE Fellow. He has been serving as Associate Editor for IEEE T-IP, T-MM, and T-CSVT as well as serving as Area Chair for ICCV, ECCV, ACM Multimedia, ICME and ICIP.
\end{IEEEbiography}

\begin{IEEEbiography}[{\includegraphics[width=1in,height=1.25in,clip,keepaspectratio]{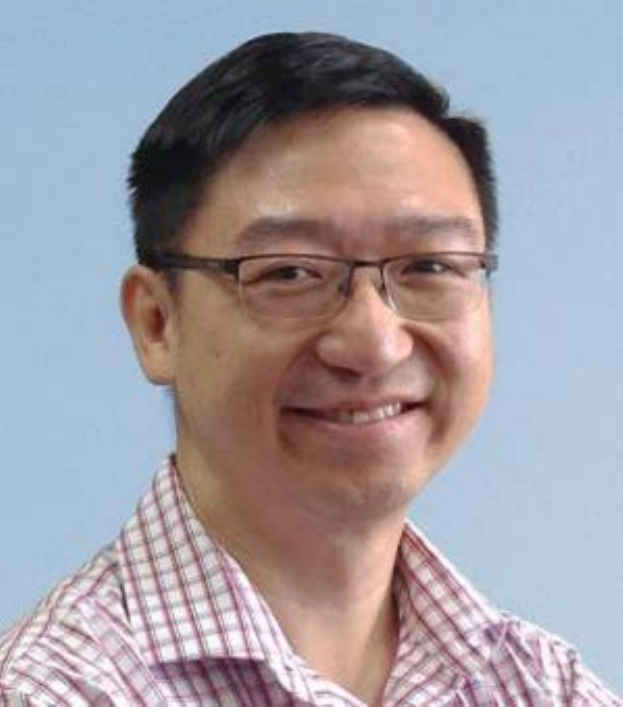}}]{Tat-Jen Cham}
received the B.A. degree in engineering
and the Ph.D. degree both from the University of Cambridge, in 1993 and 1996, respectively. He received the Jesus College Research Fellowship in Cambridge 1996-97 and was a research scientist in the DEC / Compaq Research Lab in Boston 1998-2001. He is currently an Associate Professor in the School of Computer Science and Engineering, Nanyang Technological University and a principal investigator in the NRF BeingThere Centre for 3D Telepresence. His research interests include broadly in computer vision and machine learning. He is on the editorial board of IJCV, and was a general co-chair of ACCV 2014.
\end{IEEEbiography}

\begin{IEEEbiography}[{\includegraphics[width=1in,height=1.25in,clip,keepaspectratio]{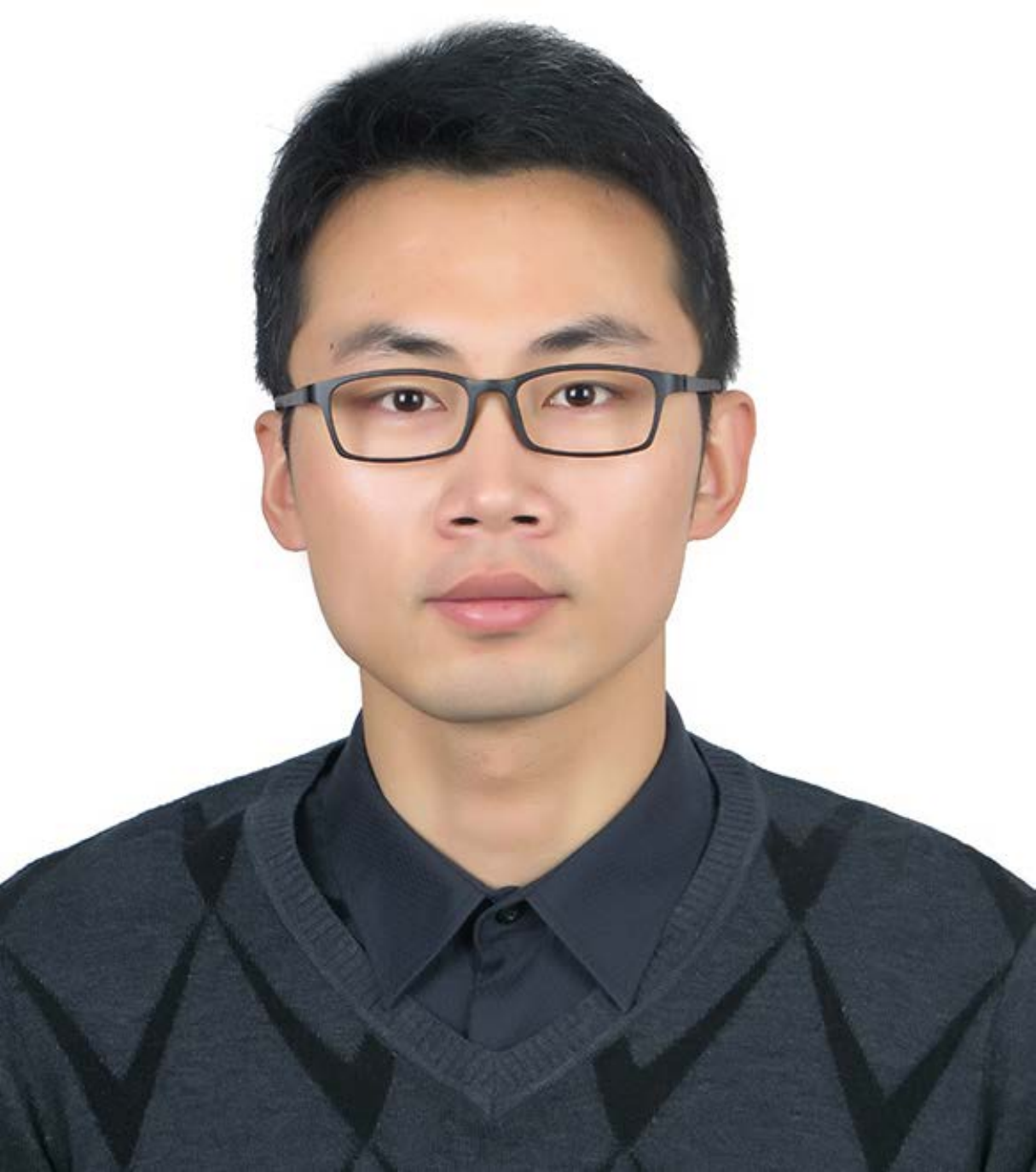}}]{Xuequan Lu}
is an Assistant Professor at the School of Information Technology, Deakin University, Australia. He has spent more than two years as a Research Fellow in Singapore. Prior to that, he received his Ph.D. degree at the Zhejiang University, China in June 2016. His research interests mainly fall into the category of visual computing, for example, geometry modeling, processing and analysis, animation/simulation, 2D data processing and analysis. He has served as a member in the International Program Committee of GMP 2021, as well as a PC member in CVM 2020 and a TPC member in ICONIP 2019. More information can be found at http://www.xuequanlu.com.
\end{IEEEbiography}

\begin{IEEEbiography}[{\includegraphics[width=1in,height=1.25in,clip,keepaspectratio]{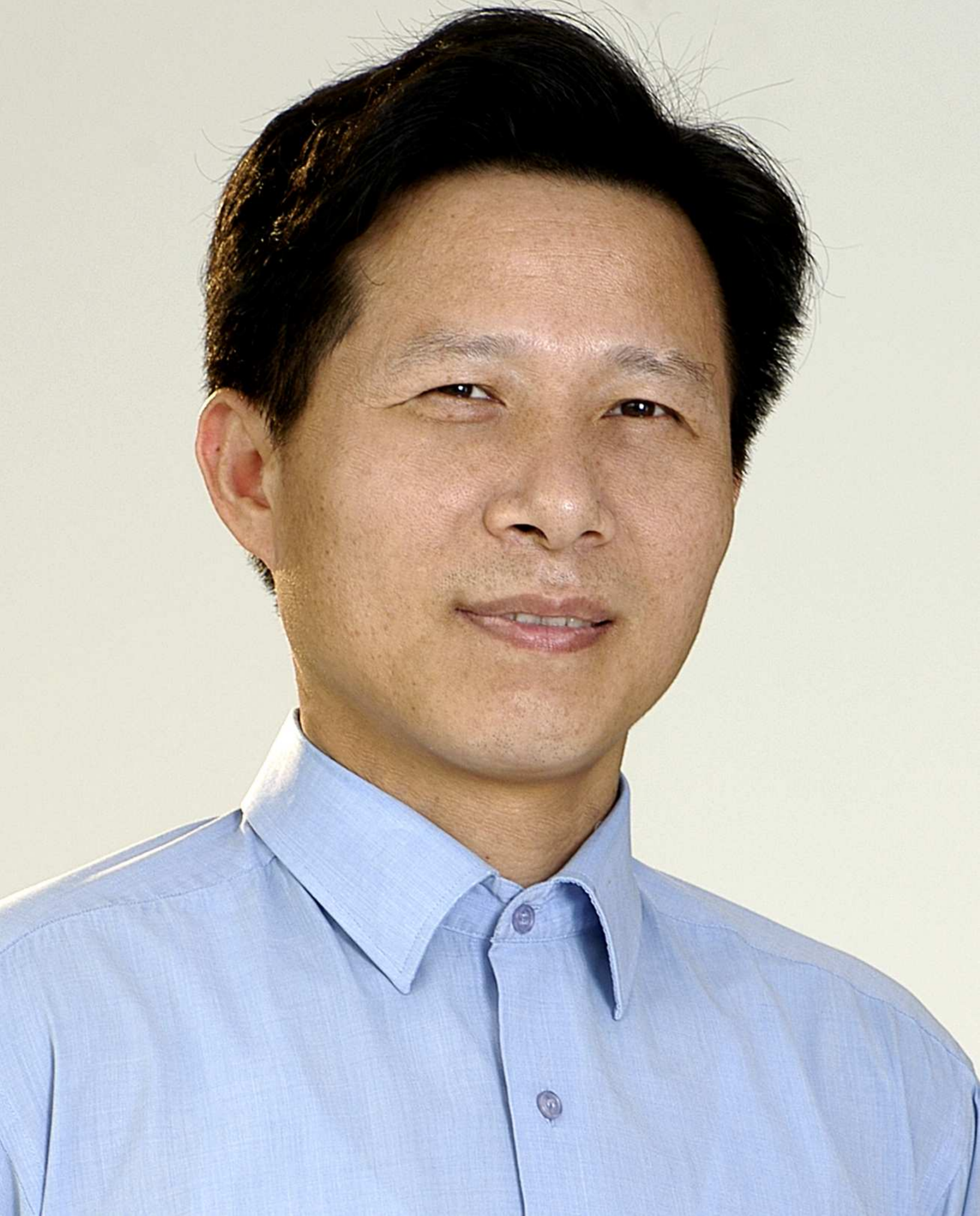}}]{Lizhuang Ma}
received his B.S. and Ph.D. degrees from the Zhejiang University, China in 1985 and 1991, respectively. He is now a Distinguished Professor, Ph.D. Tutor, and the Head of the Digital Media and Computer Vision Laboratory at the Department of Computer Science and Engineering, Shanghai Jiao Tong University, China. He was a Visiting Professor at the Frounhofer IGD, Darmstadt, Germany in 1998, and was a Visiting Professor at the Center for Advanced Media Technology, Nanyang Technological University, Singapore from 1999 to 2000. He has published more than 200 academic research papers in both domestic and international journals. His research interests include computer aided geometric design, computer graphics, scientific data visualization, computer animation, digital media technology, and theory and applications for computer graphics, CAD/CAM.
\end{IEEEbiography}


%
%
%




\end{document}